\pdfoutput=1

\documentclass[11pt]{article}

\usepackage{EMNLP2022}

\usepackage{times}
\usepackage{latexsym}

\usepackage[T1]{fontenc}

\usepackage[utf8]{inputenc}

\usepackage{microtype}

\usepackage{inconsolata}

\usepackage{amsmath, bm, amssymb, enumitem}
\usepackage{graphicx}
\newcommand{\R}{\mathbb{R}}

\title{The Geometry of Multilingual Language Model Representations}

\author{Tyler A. Chang$^{1,2}$, \quad Zhuowen Tu$^1$, \quad Benjamin K. Bergen$^1$ \\
$^1$Department of Cognitive Science \\
$^2${Halıcıoğlu} Data Science Institute \\
University of California San Diego \\
{\texttt{$\{$tachang, ztu, bkbergen$\}$@ucsd.edu}}
}

\begin{document}
\maketitle
\begin{abstract}
We assess how multilingual language models maintain a shared multilingual representation space while still encoding language-sensitive information in each language. Using XLM-R as a case study, we show that languages occupy similar linear subspaces after mean-centering, evaluated based on causal effects on language modeling performance and direct comparisons between subspaces for 88 languages. The subspace means differ along language-sensitive axes that are relatively stable throughout middle layers, and these axes encode information such as token vocabularies. Shifting representations by language means is sufficient to induce token predictions in different languages. However, we also identify stable language-neutral axes that encode information such as token positions and part-of-speech. We visualize representations projected onto language-sensitive and language-neutral axes, identifying language family and part-of-speech clusters, along with spirals, toruses, and curves representing token position information. These results demonstrate that multilingual language models encode information along orthogonal language-sensitive and language-neutral axes, allowing the models to extract a variety of features for downstream tasks and cross-lingual transfer learning.
\end{abstract}

\section{Introduction}
Despite state-of-the-art performance on a wide variety of multilingual NLP tasks \citep{conneau-etal-2020-unsupervised,hu-etal-2020-extreme,liang-etal-2020-xglue,lin-etal-2021-few,xue-etal-2021-mt5}, the internal structure of multilingual language model representation spaces is not well understood.
The success of these models is often attributed to a shared multilingual space across languages, with claims that the models ``map representations coming from different languages in a single shared embedding space'' \citep{conneau-etal-2020-emerging} and that ``different languages are close to a shared space'' \citep{pires-etal-2019-multilingual}.
While intuitive, many of these assumptions have not been examined in detail.
For example, the models have been shown to retain language-sensitive information such as linguistic typological features (language-sensitive representations; \citealp{choenni-shutova-2020-what, liang-etal-2021-locating, rama-etal-2020-probing}) despite also exhibiting cross-lingual alignment of representations (language-neutral representations; \citealp{conneau-etal-2020-emerging,libovicky-etal-2020-language,pires-etal-2019-multilingual}).
It remains unclear how the underlying geometry of the representation space facilitates this integration of language-sensitive and language-neutral information, and the structure of the space with respect to individual features and axes remains an open question.

\setlength{\belowcaptionskip}{-0.1cm}
\begin{figure}[t]
    \centering
    \includegraphics[width=7.5cm]{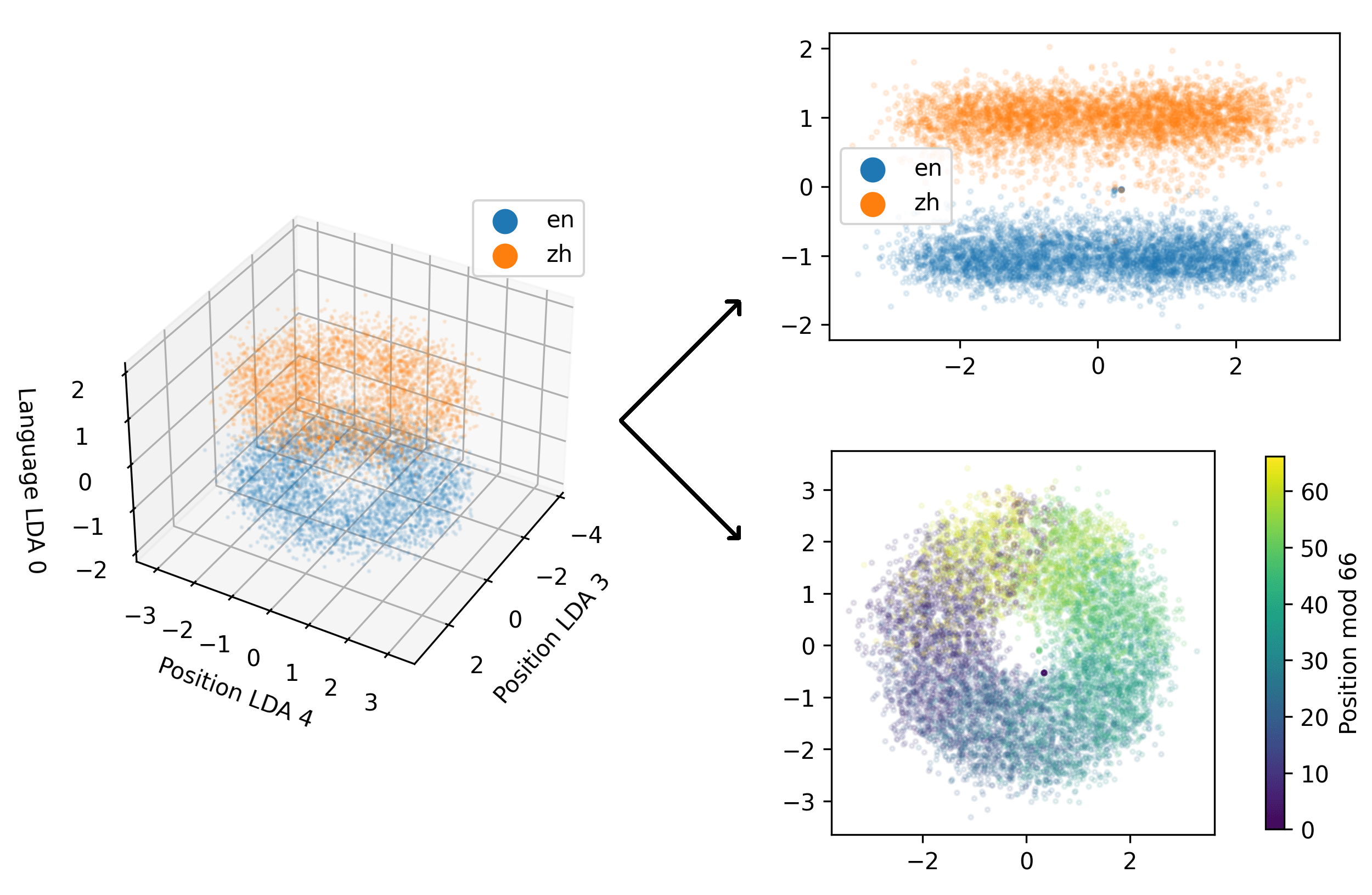}
    \caption{XLM-R representations in layer six projected onto a linear subspace where two axes are language-neutral (horizontal axes), and one axis is language-sensitive (vertical axis).
    Projecting from the side visualizes the language-sensitive axis (top right). Projecting from the top down visualizes the language-neutral axes encoding token position information, forming a nearly perfect torus in the representation space (bottom right).}
    \label{fig:projections-lang-positions}
\end{figure}
\setlength{\belowcaptionskip}{0.0cm}

In this work, we present an analysis of the geometry of multilingual language model representations using downstream language modeling predictions, direct comparisons between language subspaces, and visualizations of representations projected onto low-dimensional subspaces.
Using the multilingual language model XLM-R as a case study \citep{conneau-etal-2020-unsupervised}, we find that language subspaces are similar to one another after mean-centering, and differences between subspace means encode language-sensitive information such as language vocabularies.
We identify language-sensitive axes that cluster representations by language family, and we identify language-neutral axes that encode token positions and part-of-speech.
These axes remain relatively stable across layers, suggesting that multilingual language models maintain stable substructures for individual features during processing.
Our results highlight the importance of representational geometry in understanding multilingual language model representations, laying the groundwork for future research in multilingual subspace geometry and interpretable multilingual learning.\footnote{Code is available at \url{https://github.com/tylerachang/multilingual-geometry}.}

\section{Related work}
Previous work has considered how multilingual language models encode different types of information.
For example, mean representation distances between languages correlate with phylogenetic distances between languages \citep{rama-etal-2020-probing}, and individual representations can be used to predict linguistic typological features \citep{choenni-shutova-2020-what}, particularly after projecting onto language-sensitive subspaces \citep{liang-etal-2021-locating}.
The models also maintain language-neutral subspaces that encode information that is shared across languages.
Syntactic information is encoded largely within a shared syntactic subspace \citep{chi-etal-2020-finding}, token frequencies may be encoded similarly across languages \citep{rajaee-pilehvar-2022-isotropy}, and representations shifted according to language means facilitate cross-lingual parallel sentence retrieval \citep{libovicky-etal-2020-language, pires-etal-2019-multilingual}.
However, these studies have primarily focused on specific features or probing tasks, with less focus on how the broader geometry of the original representation space facilitates feature encodings and interactions.
To better understand this broader context, we consider overall language subspaces in multilingual language models, identifying axes that encode language-sensitive and language-neutral features.

\section{Language subspaces}
\label{sec:language-subspaces}
As an initial assessment of the geometry of multilingual representation spaces, we identified affine (i.e. mean-shifted linear) subspaces for 88 languages in the multilingual language model XLM-R \citep{conneau-etal-2020-unsupervised} using singular value decomposition (SVD) over contextualized token representations in each language.
Previous work has considered SVD primarily in the context of singular value canonical correlation analysis (SVCCA; \citealp{raghu-etal-2017-svcca}) to quantify informational similarity between sets of representations (e.g. \citealp{kudugunta-etal-2019-investigating,saphra-lopez-2019-understanding}), without considering the geometry of the un-transformed subspaces. 
Here, we found that SVD identifies affine subspaces that sufficiently account for language modeling performance in individual languages, and the subspaces are similar to one another after subtracting language means, particularly in middle layers.

\subsection{Model and dataset}
In all experiments, we used the pre-trained language model XLM-R, which has achieved state-of-the-art performance on a variety of multilingual NLP tasks \citep{conneau-etal-2020-unsupervised}.
XLM-R follows the Transformer architecture of BERT and RoBERTa \citep{devlin-etal-2019-bert, liu-etal-2019-roberta}, but the model is trained to predict masked tokens in 100 languages.
To extract contextualized token representations from XLM-R, we inputted text sequences from the OSCAR corpus of cleaned web text data \citep{abadji-etal-2021-ungoliant}, concatenating consecutive sentences such that each sequence contained 512 tokens.
We used the vector representations outputted by each Transformer layer as our token representations in layers one through twelve, and we used the uncontextualized token embeddings for layer zero.
We considered the 88 languages that appear in both the XLM-R pre-training corpus and the OSCAR corpus, excluding languages with fewer than 100 sequences in OSCAR.\footnote{Experimental details are outlined in Appendix \ref{app:experimental-details}.}

\subsection{Affine language subspaces}
\label{sec:affine-subspaces}
We defined an affine subspace for each language $A$ using the language's mean representation $\bm{\mu}_A \in \R^d$ along with $k$ directions of maximal variance in the language, defined by an orthonormal basis $\bm{V}_A \in \R^{d \times k}$.
To identify this subspace, we applied singular value decomposition (SVD) centered at $\bm{\mu}_A$ using 262K contextualized token representations from language $A$ (512 sequences in the OSCAR corpus).
We selected the subspace dimensionality $k$ such that the subspace accounted for $90\%$ of the total variance in the language.\footnote{
Results were qualitatively similar for subspaces accounting for variance proportions in [75\%, 90\%, 95\%, 99\%].}
Across all layers, the median subspace dimensionality was 335, less than half of the original 768 dimensions.

\setlength{\belowcaptionskip}{-0.3cm}
\begin{figure}[t]
    \centering
    \includegraphics[width=7.5cm]{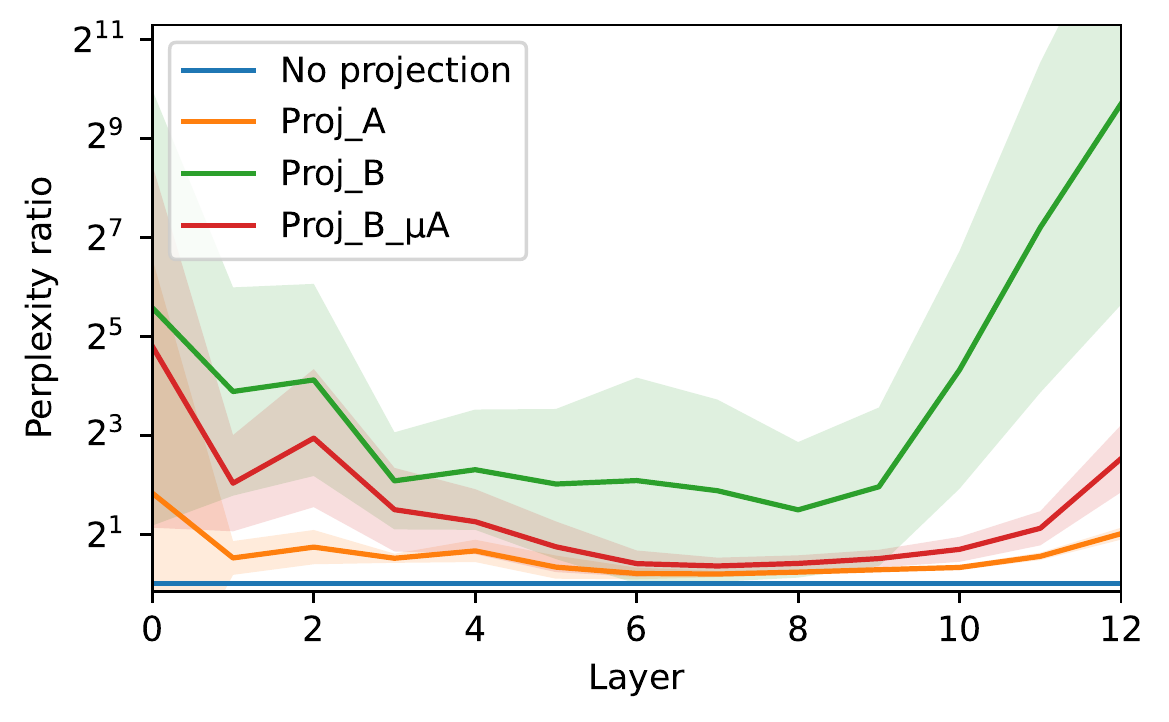}
    \caption{Language modeling perplexity scores in language $A$ when projecting representations onto different affine language subspaces in each layer.
    We report the mean perplexity ratio (the projected perplexity score divided by the original perplexity score) over all 88 evaluation languages $A$.
    For projections onto language $B$ subspaces, we report the mean perplexity ratio for all pairs where the evaluation language $A$ did not match the subspace language $B$.
    Higher values indicate worse language modeling performance.
    Due to the interpretation of perplexities as inverse probabilities, we computed geometric rather than arithmetic means.
    Shaded regions indicate one (geometric) standard deviation from the mean.
    }
    \label{fig:perplexity-comparison}
\end{figure}
\setlength{\belowcaptionskip}{0.0cm}

\textbf{Affine subspaces accounted for language modeling performance.}
To assess the extent to which affine subspaces encoded relevant information in their corresponding languages, we evaluated language modeling perplexity scores for each language $A$ when projecting representations onto the corresponding language subspace:
\[ \textrm{Proj}_A (\bm{x}) = \bm{V}_A \bm{V}_A^T (\bm{x} - \bm{\mu}_A) + \bm{\mu}_A  \]
We computed the ratio of the projected perplexity to the original perplexity in language $A$.
As shown in Figure \ref{fig:perplexity-comparison} ($\textrm{Proj}_A$), there were generally only minor increases in perplexity despite projecting onto subspaces with less than half the dimensionality of the original space.
This suggests that affine language subspaces encode much of the information relevant to the language modeling task in their corresponding languages.

\textbf{Language subspaces differed from one another.}
To evaluate whether the model was simply using the same affine subspace for all languages, we evaluated perplexities in each language $A$ when projecting onto subspaces for different languages $B$.
As shown in Figure \ref{fig:perplexity-comparison} ($\textrm{Proj}_B$), perplexities increased substantially when projecting onto these other language subspaces, suggesting that the model maps text in different languages into distinct subspaces.
However, we note that these projections projected onto subspaces passing through each $\bm{\mu}_B$, which may have been quite far from the original mean of the evaluation language $\bm{\mu}_A$.
Thus, the high perplexities when projecting onto other language subspaces may simply have been a result of projecting onto subspaces passing through $\bm{\mu}_B$ rather than $\bm{\mu}_A$.

\textbf{Mean-shifted subspaces were similar to one another.}
We again evaluated perplexities in each language $A$ projected onto languages $B$, but we shifted the language $B$ subspaces such that they passed through $\bm{\mu}_A$.\footnote{Note that $\textrm{Proj}_{B, \bm{\mu}_A} (\bm{x}) = \bm{V}_B \bm{V}_B^T (\bm{x} - \bm{\mu}_A) + \bm{\mu}_A$ is equal to $\textrm{Proj}_B(\bm{x})$ shifted by the constant vector $(\bm{I} - \bm{V}_B \bm{V}_B^T)(\bm{\mu}_A - \bm{\mu}_B)$. In other words, the projected representation is shifted by the component of $\bm{\mu}_A - \bm{\mu}_B$ that is perpendicular to the language $B$ subspace.}
As shown in Figure \ref{fig:perplexity-comparison} ($\textrm{Proj}_{B, \bm{\mu}_A}$), particularly for middle layers, perplexities under these mean-shifted projections were only moderately higher than when projecting onto the language $A$ subspace.
In other words, projecting onto the affine subspace for language $B$ shifted onto $\bm{\mu}_A$ was similar to projecting onto the affine subspace for language $A$ itself (by default passing through $\bm{\mu}_A$).
This suggests that in middle layers, the affine subspaces for different languages are similar to one another when shifted according to language means.
In further support of this finding, we note that the perplexity gap between $\textrm{Proj}_B$ and $\textrm{Proj}_{B, \bm{\mu}_A}$ tended to widen in deeper layers.
This implies that the language modeling performance degradation caused by language projection $\textrm{Proj}_B$ can be accounted for largely by language mean in deeper layers.
In these deeper layers, it appears that the differences between language subspaces compress into differences between subspace means.

\subsection{Subspace distances}
\label{sec:subspace-distances}
As a complementary metric, we directly quantified distances between the subspaces themselves.
To do this, we first note that our SVD approach to computing affine language subspaces identifies principal axes with corresponding variances for each language.
By interpreting these axes and variances as covariance matrices (instead of using them to define affine subspaces), we can adopt theoretically-motivated distance metrics from mathematics that quantify distances between positive definite matrices \citep{bonnabel-sepulchre-2009-riemannian}.\footnote{A positive definite matrix is a symmetric matrix with positive eigenvalues. Our selected distance metric is described in more detail in Appendix \ref{app:distance-metric} and in \citet{bonnabel-sepulchre-2009-riemannian}.}
Specifically, we define the distance between two positive definite matrices $\bm{K}_A, \bm{K}_B \in \R^{d \times d}$ as:
\begin{equation}
\label{eq:distance}
\textrm{Distance}(\bm{K}_A, \bm{K}_B) = \sqrt{\sum_i \log^2(\lambda_i)}
\end{equation}
where $\lambda_i$ are the $d$ positive real eigenvalues of $\bm{K}_A^{-1} \bm{K}_B$ \citep{bonnabel-sepulchre-2009-riemannian}.
This distance metric is both symmetric and invariant to linear transformations (e.g. rotations, reflections, and scaling).
However, because the metric ignores subspace means, it can only be considered as a distance metric between mean-centered subspaces.

We computed pairwise distances between the 88 language subspaces identified in the previous section.
To gain an intuitive understanding of the subspace distances, we compared the true distances to the distances between subspaces rotated by $\theta$ degrees or scaled by a multiplier $\gamma$ along each axis.
For example, the distance between the Spanish and Chinese subspaces in layer eight was approximately equal to the mean distance from a language subspace to itself before and after rotation by two degrees or scaling by 1.53x.\footnote{A rotation was defined as a $\theta$-degree rotation in each of the (randomly selected) $d // 2$ independent planes of rotation. A scaling was defined as multiplication or division by $\gamma$ along each of the $d$ axes of variance.
Details for the rotation and scaling comparisons can be found in Appendix \ref{app:rotations-scalings}.}
This way, we were able to consider the distance between any two subspaces in terms of the analogous rotation or scaling.

\setlength{\belowcaptionskip}{-0.3cm}
\begin{figure}[t]
    \centering
    \includegraphics[width=7.5cm]{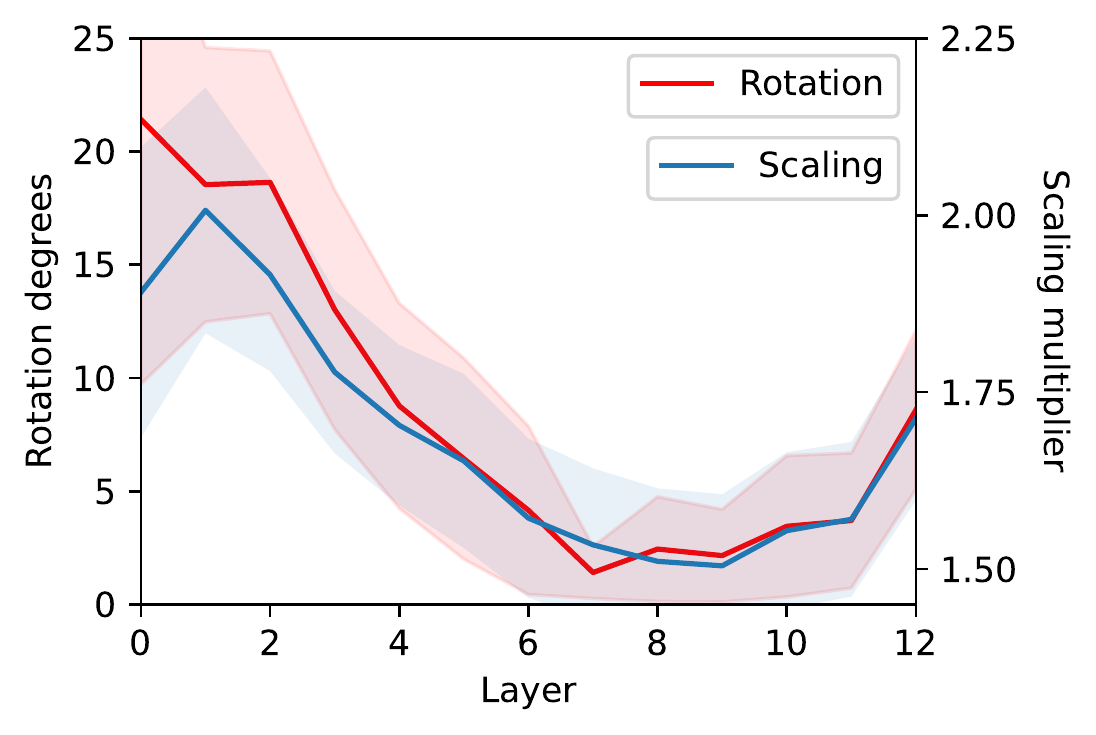}
    \caption{Analogous rotations and scalings between mean-centered language subspaces in each layer.
    Shaded regions indicate one standard deviation from the mean. In middle and late-middle layers, mean-centered subspaces were more similar to one another.
    }
    \label{fig:rotations-scalings}
\end{figure}
\setlength{\belowcaptionskip}{0.0cm}

\textbf{Again, mean-shifted subspaces were similar to one another.}
The mean rotation degrees and scaling multipliers for language subspace distances in each layer are shown in Figure \ref{fig:rotations-scalings}.
In line with the perplexity comparison results in Section \ref{sec:affine-subspaces}, language subspaces in middle layers were surprisingly similar to one another after mean-centering.
In layers six through eleven, the mean subspace distances were equivalent to subspace rotations by less than five degrees and subspace scalings by less than 1.6x.
This result aligns with previous work suggesting that representations from middle layers in multilingual language models are often the most cross-linguistically aligned, particularly after adjusting for language means \citep{libovicky-etal-2020-language, pires-etal-2019-multilingual}.
In these middle layers, the model representations are further from both the original input and the final language modeling prediction, both of which are highly language-sensitive.

\section{Language-sensitive axes}
\label{sec:language-sensitive-axes}
Thus, it appears that languages occupy similar subspaces in multilingual language models after mean-centering, based on converging evidence from downstream language modeling performance and direct comparisons between language subspaces.
However, differences in subspace means demonstrate that the language subspaces still differ along particular axes.
Intuitively, these axes should encode language-sensitive information, information that has high mutual information with the input language identity.
For example, the word order or specific tokens present in a raw linguistic input are highly informative of the input language.
In this section, we consider whether axes connecting subspace means encode language-sensitive features such as token vocabularies.
Indeed, we found that shifting representations by language means was sufficient to induce language modeling predictions in arbitrary target languages.
We then used linear discriminant analysis (LDA) to explicitly identify language-sensitive axes, finding that these axes were surprisingly stable across middle layers.

\subsection{Inducing target language vocabulary}
\label{sec:inducing-vocabulary}
We assessed whether shifting and projecting representations according to language means and subspaces was sufficient to induce language modeling predictions in different languages.
We defined a language's vocabulary as the set of tokens with frequency at least 1e-6 (one occurrence per million tokens) in up to one billion tokens from the OSCAR corpus for that language.
In XLM-R, these vocabularies ranged from 3K to 24K tokens from the original 250K token vocabulary.
From each vocabulary, we excluded a set of common tokens (945 tokens) that appeared in at least 90\% of the 88 languages; these common tokens consisted primarily of punctuation, numbers, and one- to two-character sequences of Latin characters.

\begin{figure}[t]
    \centering
    \includegraphics[width=7.5cm]{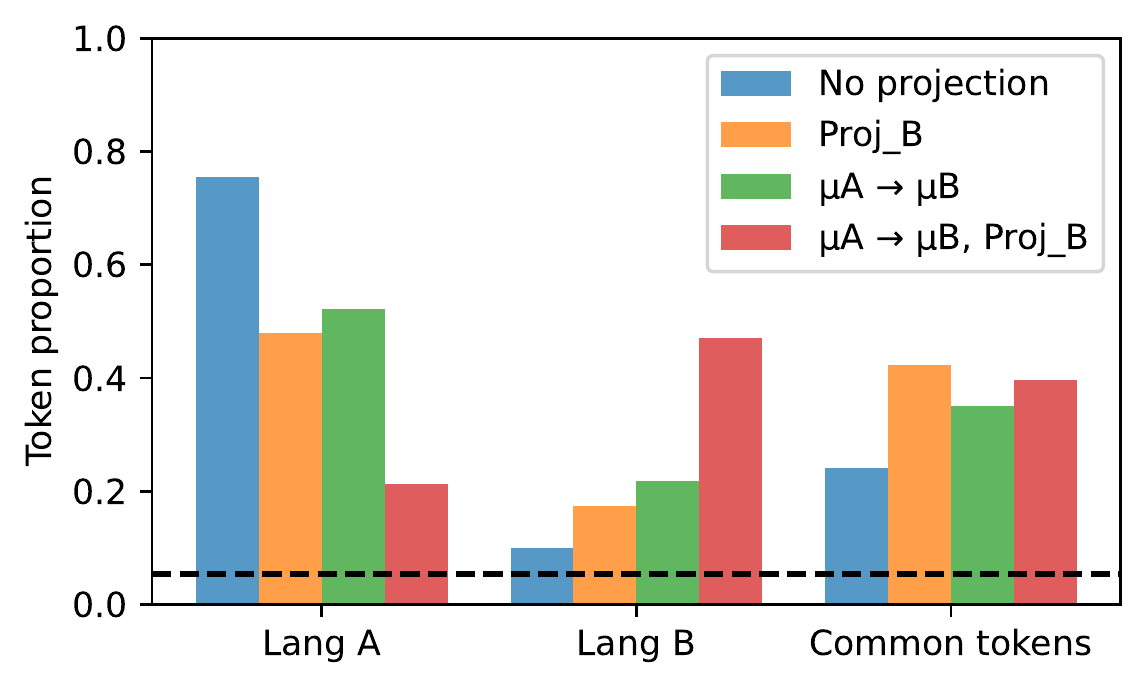}
    \caption{The proportion of tokens in the original evaluation language $A$, the target language $B$, and in a set of common tokens across languages (e.g. numerals and punctuation), after different types of mean-shifting and language subspace projections in the language model. Layers were projected individually, and results were averaged across layers. Results varied substantially based on the evaluation and target languages (standard deviations in all conditions between 0.11 and 0.25), likely due to differences in vocabulary overlap between language pairs, raw vocabulary sizes, and model biases towards certain languages \citep{wu-dredze-2020-languages}. The dashed line indicates the average proportion of tokens (5.4\%) that would be in any given language if predicting based on random chance.
    }
    \label{fig:token-proportions}
\end{figure}

We collected language modeling predictions for 512 sequences in each evaluation language $A$, computing the proportion of predicted tokens in the language $A$ vocabulary.
For all languages, nearly all predicted tokens were either common tokens or tokens in the evaluation language ($M = 99.5\%, SD=0.2\%$).\footnote{Because nearly all predicted tokens in the unmodified language model were either in the evaluation language $A$ or common tokens, the proportion of tokens in other languages $B$ in this condition were essentially due to vocabulary overlap between languages $A$ and $B$.}
The mean proportions of common tokens and tokens in the evaluation language $A$ are shown in Figure \ref{fig:token-proportions}.

\setlength{\belowcaptionskip}{-0.3cm}
\setlength{\belowcaptionskip}{0.0cm}

\textbf{Shifting by language means induced target language vocabulary.}
We considered the same token proportions when representations in language $A$ were shifted towards each possible target language $B$ by adding the shift vector $\bm{\mu}_B - \bm{\mu}_A$.
As shown in Figure \ref{fig:token-proportions}, shifting towards $\bm{\mu}_B$ substantially increased the proportion of predicted tokens in language $B$, decreasing the proportion of predicted tokens in language $A$.
This suggests that the model encodes token vocabularies to some degree along axes where language means differ.

\textbf{Projecting onto subspaces induced additional target language vocabulary.}
Next, we considered whether projecting onto the language $B$ subspace induced additional predictions in language $B$.
Projecting onto this subspace is equivalent to removing information along axes orthogonal to the language $B$ subspace, by simply setting the representation equal to $\bm{\mu}_B$ along these axes.
As shown in Figure \ref{fig:token-proportions}, this projection alone induced a proportion of target language tokens similar to mean-shifting.
This suggests that along some axes, the language mean alone captures relevant information about language $B$ vocabularies.

Then, we considered the combination of mean-shifting and subspace projection; geometrically, this set axes orthogonal to the language $B$ subspace equal to $\bm{\mu}_B$, and it shifted representations according to $\bm{\mu}_B - \bm{\mu}_A$ along axes within the language $B$ subspace.
Indeed, this transformation further increased the proportion of predicted tokens in language $B$, beyond mean-shifting and projection individually (Figure \ref{fig:token-proportions}).
Compared to the unmodified language model, this transformation induced 4.7x more predicted tokens in the target language ($10\% \to 47\%$), and 3.5x fewer predicted tokens in the original evaluation language ($75\% \to 21\%$).
These results suggest that along some axes, language-sensitive information can be captured simply by language means (setting values equal to $\bm{\mu}_B$); along other axes, representation distributions from other languages $A$ can be shifted towards language $B$ (shifting values by $\bm{\mu}_B - \bm{\mu}_A$).
A more detailed interpretation of types of language-sensitive axes is included in Appendix \ref{app:language-sensitive-neutral}.

\setlength{\belowcaptionskip}{0.0cm}
\begin{figure*}[t]
    \centering
    \includegraphics[width=13.1cm]{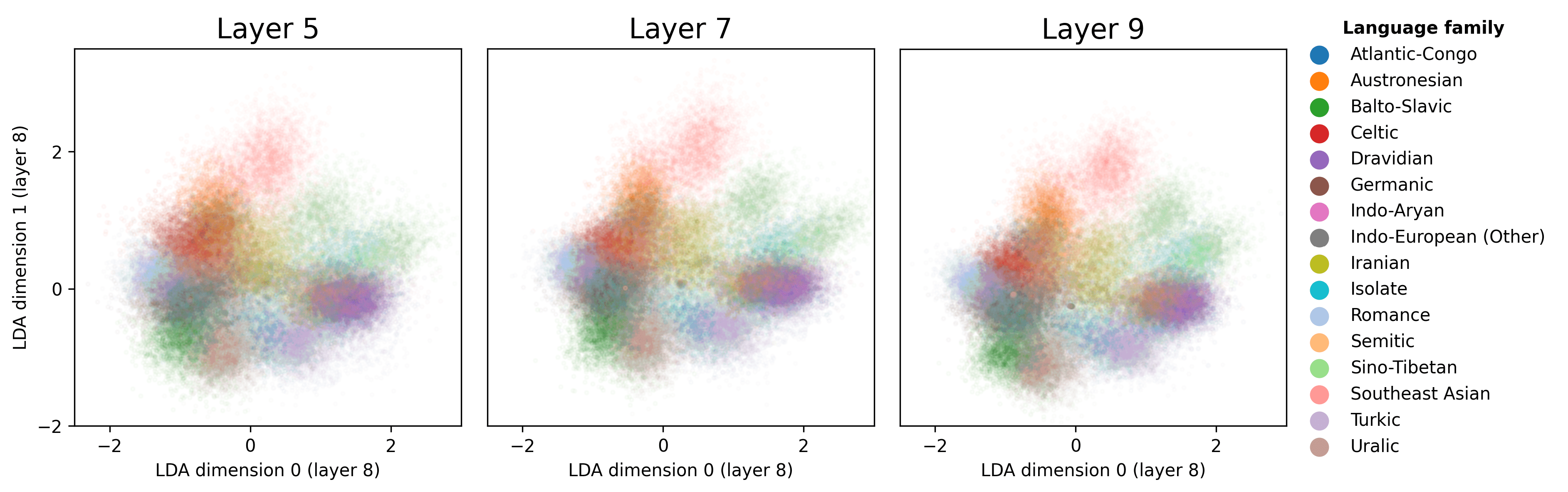}
    \caption{Representations in layers five, seven, and nine projected linearly onto the first two LDA axes that separate languages in layer eight. Representations remained relatively unchanged along these axes as they passed through middle layers of the model. Detailed plots are included in Appendix \ref{app:language-lda-plots}.
    }
    \label{fig:lang-lda}
\end{figure*}
\setlength{\belowcaptionskip}{0.0cm}

\subsection{Linear discriminant analysis (LDA)}
\label{sec:language-lda}
In the previous section, we showed that language subspaces differ along language-sensitive axes (e.g. axes connecting language means), and these axes encode information such as token vocabularies.
Next, as in \citet{liang-etal-2021-locating}, we applied linear discriminant analysis (LDA) to identify specific axes that separate language subspaces.
Given $n$ sets of representations (in this case, one set of 4K randomly sampled representations for each language), LDA computes $n-1$ axes that maximize separation between the sets.
Building upon \citet{liang-etal-2021-locating}, we directly visualized representations projected onto the identified language-sensitive axes.

\textbf{Languages clustered by family.}
When representations were projected onto the first axes identified by LDA, they clustered loosely by language family (Figure \ref{fig:lang-lda}).
This aligns with the findings of \citet{liang-etal-2021-locating}, who found that LDA axes encode linguistic typological features and language families.
In Appendix \ref{app:language-lda-plots}, we include LDA plots for each layer, including all 88 individual language means.
In earlier layers and the final hidden layer, representations appeared to cluster more by script; indeed, these layers are closer to either the original token inputs or the output token predictions.

\textbf{Language-sensitive axes were stable in middle layers.}
Notably, the axes identified by LDA resulted in surprisingly similar projections for middle layers (e.g. layers five through nine).
In fact, for these layers, we were able to project representations onto the same language-sensitive axes regardless of layer.
For example, all plots in Figure \ref{fig:lang-lda} project onto the language-sensitive axes computed specifically for layer eight, but they result in nearly identical projections for representations in layers five through nine.
Projections onto these axes for the remaining layers are shown in Appendix \ref{app:language-lda-plots}, remaining similar for layers three through eleven.
We observed qualitatively similar results for the first ten language-sensitive axes identified by LDA.
These results suggest that representations remain largely unchanged along language-sensitive axes as they pass through middle and late-middle layers.
In these layers, the language model may be processing more semantic information \citep{jawahar-etal-2019-bert,tenney-etal-2019-bert}, transforming representations along more language-neutral axes.

\section{Language-neutral axes}
\label{sec:language-neutral-axes}
It is then apparent that multilingual language models encode language-sensitive information along language-sensitive axes.
However, we still do not have a clear picture of how information is encoded along presumed language-neutral axes.
Motivated by the use of LDA to identify language-sensitive axes, we used LDA to identify axes that encode potentially more language-neutral information: token positions and part-of-speech (POS).
We assessed whether the identified axes encoded the corresponding features in language-neutral ways.

\setlength{\belowcaptionskip}{-0.3cm}
\begin{figure}[t]
    \centering
    \includegraphics[width=7.5cm]{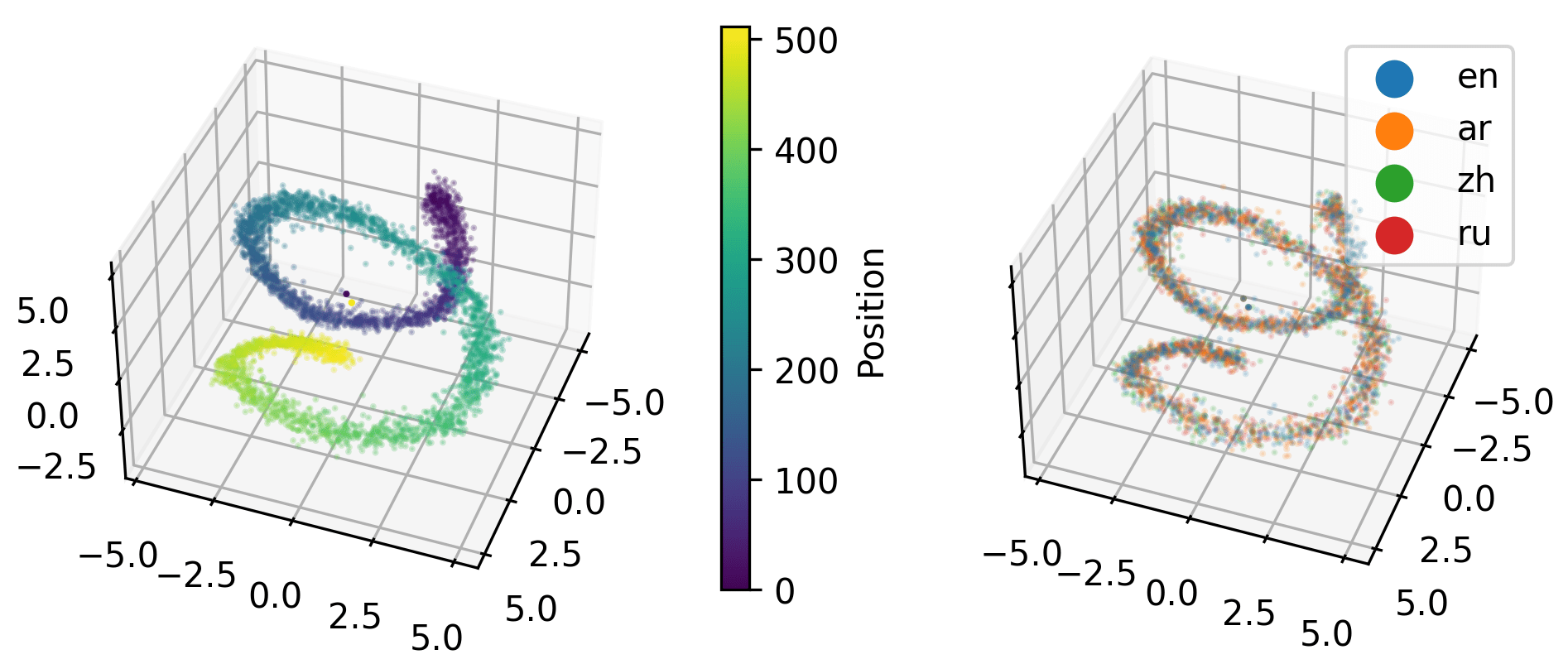}
    \caption{Representations in layer eight projected linearly onto the first three LDA axes that separate different token positions. As shown in the right image, token position information was primarily encoded independently of input language.
    }
    \label{fig:position-lda}
\end{figure}
\setlength{\belowcaptionskip}{0.0cm}

\subsection{Token positions}
\label{sec:position-axes}
First, we identified axes that encode tokens' positions in input sequences.
Notably, token positions are encoded language-neutrally as absolute position embeddings before the first Transformer layer in XLM-R.
Unless the model transforms position information in a language-sensitive way, the information will remain encoded language-neutrally.
Still, identifying token position axes serves to verify our assumptions about the language neutrality of position information in the model and to better understand how this information is represented.

\textbf{Position axes were language-neutral.}
We performed LDA on sets of representations corresponding to every sixteen token positions, identifying axes that separated the different positions.
We used 8K representations sampled uniformly from all languages for each position index.
We projected representations from all token positions onto the identified position axes to qualitatively determine whether the axes encoded position information language-neutrally.
Indeed, as shown in Figure \ref{fig:position-lda}, along the position axes, we could qualitatively identify a token's position in the input sequence without knowing its source language, demonstrating that token position information remains largely language-neutral as it passes through the model.

\setlength{\belowcaptionskip}{-0.3cm}
\begin{figure}[t]
    \centering
    \includegraphics[width=7.5cm]{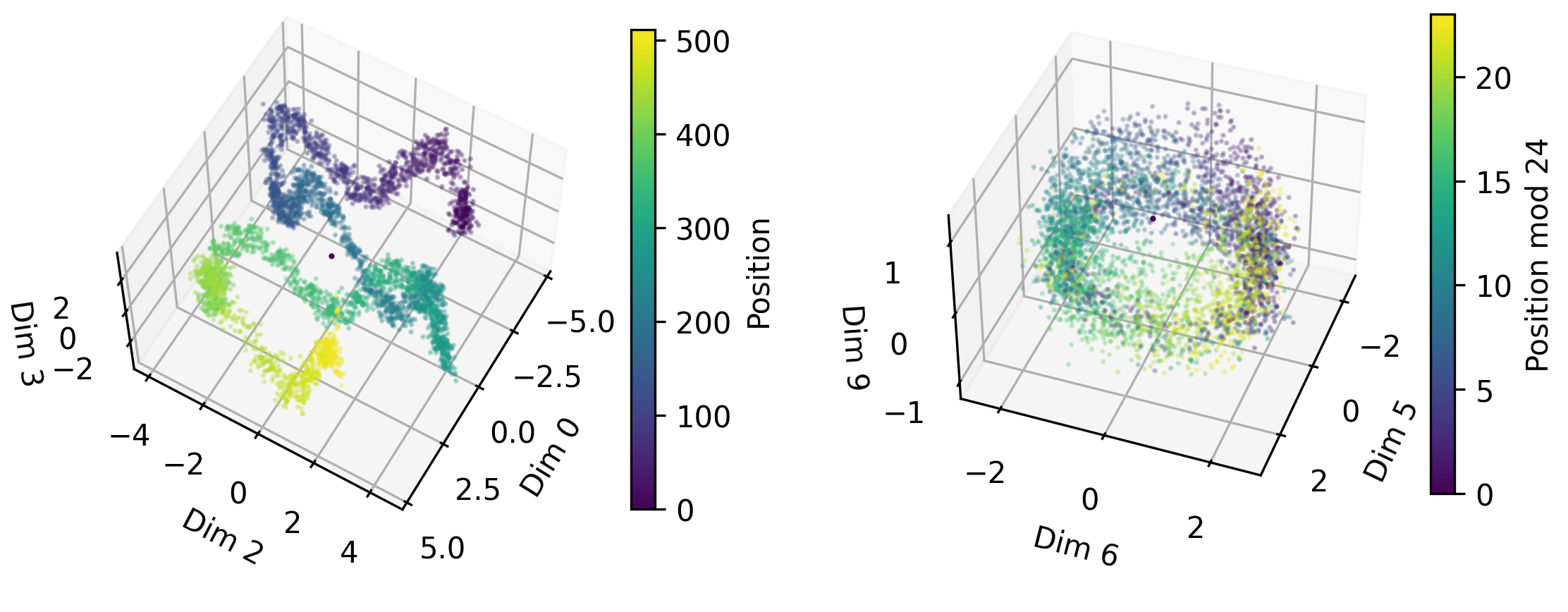}
    \caption{Representations in layer four projected linearly onto LDA axes that separate different token positions. In some subspaces, curves encoded absolute positions (left), while in others, toruses encoded relative positions (right).
    }
    \label{fig:position-lda-curves}
\end{figure}
\setlength{\belowcaptionskip}{0.0cm}

\textbf{Position information was encoded along nonlinear structures.}
As shown in Figure \ref{fig:position-lda-curves}, token position information was encoded along toruses, spirals, and curves in the position subspaces.
These structures were similar to the ``Swiss-Roll manifold'' identified along directions of maximal variance in unidirectional monolingual language models in \citet{cai-etal-2021-isotropy}.\footnote{Indeed, we were able to identify spirals encoding token position information in middle layers simply by projecting onto directions of maximal variance (within each language or across all languages) in the multilingual model. This suggests that token position information is often encoded along directions of high variance in language model representations.}
We hypothesize that this spiral structure may be due to the need for both relative and absolute position information in the model.
In some cases, it is useful to know a token's relative position to nearby tokens; this information can be encoded through toruses, whose circular structure can encode token positions modulo some window size (e.g. Figure \ref{fig:position-lda-curves}, right).
Then, angles on the torus represent relative position distances.
In other cases, it may be useful to know a token's absolute position in the overall sequence; this can be encoded along a single curve or linear dimension (e.g. Figure \ref{fig:position-lda-curves}, left).
Combining this relative and absolute position information can be attained through multidimensional spirals (e.g. the 3D spiral in Figure \ref{fig:position-lda}), which project onto lines and curves along some axes and toruses along others.
Future work may investigate how these nonlinear structures arise from and interact with the dot product self-attention mechanism in Transformer models.\footnote{Dot products are proportional to angle cosines on toruses. If angles encode relative token positions (e.g. Figure \ref{fig:position-lda-curves}, right), the self-attention mechanism has easy access to relative position information.}

\textbf{Position representations were stable across layers.}
Similar to the projections onto language-sensitive axes in Section \ref{sec:language-lda}, we were able to project representations onto token position axes identified for other layers while still producing nearly identical curves to those in Figure \ref{fig:position-lda} (see Appendix \ref{app:position-lda-plots} for plots).
In fact, we found that these axes were stable as early as the second hidden layer, remaining largely unchanged until the last hidden layer.
This suggests that each layer applies only minimal transformations to the representations along language-neutral position axes, leaving internal position representations largely unchanged as they pass through the model.

\setlength{\belowcaptionskip}{-0.3cm}
\begin{figure}[t]
    \centering
    \includegraphics[width=6.5cm]{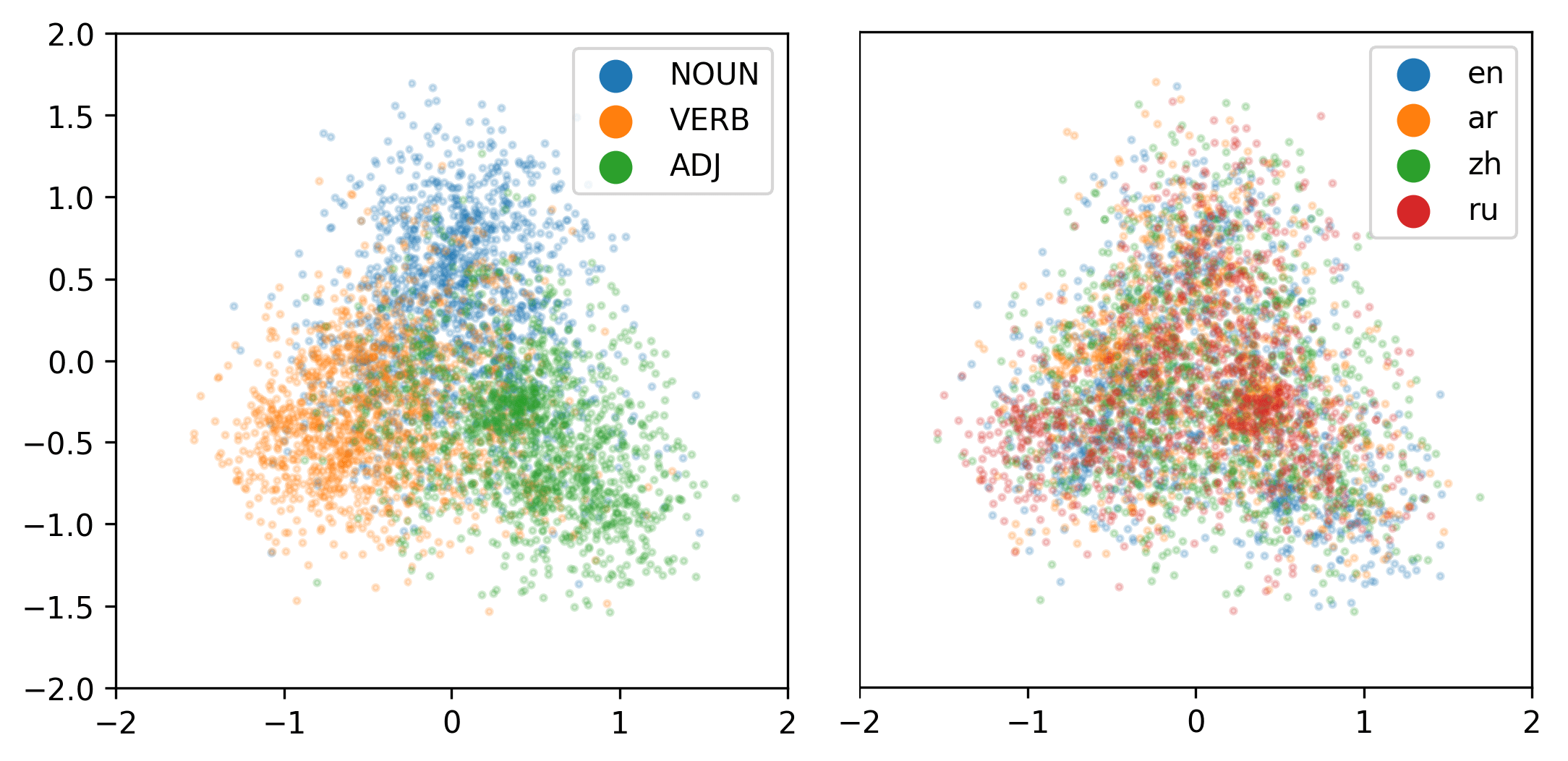}
    \caption{Representations in layer two projected linearly onto LDA axes that separate nouns, verbs, and adjectives. As shown in the right image, representations clustered independently of input language along these axes.
    Additional plots are included in Appendix \ref{app:pos-lda-plots}.
    }
    \label{fig:pos-lda}
\end{figure}
\setlength{\belowcaptionskip}{0.0cm}

\subsection{Part-of-speech}
\label{sec:pos-axes}
As a stronger test of whether multilingual language models align language-neutral information along language-neutral axes, we considered axes that encode tokens' part-of-speech (POS).
Unlike token positions, POS is not inputted directly into the model; in order to encode POS in a language-neutral way, the model must align features (e.g. features of nouns vs. verbs) cross-linguistically without supervision.

\setlength{\belowcaptionskip}{-0.3cm}
\begin{figure}[t]
    \centering
    \includegraphics[width=7.5cm]{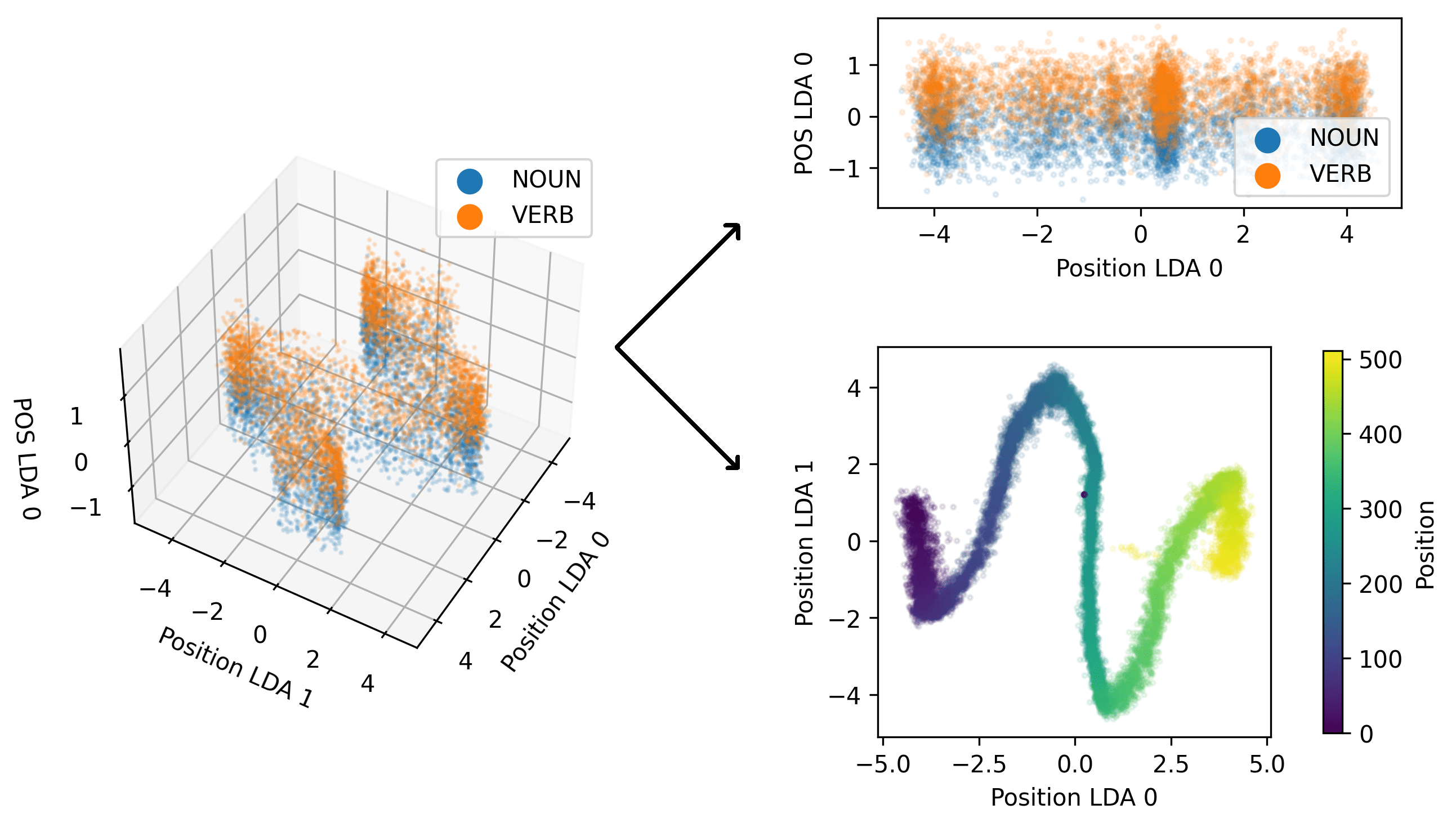}
    \caption{Representations from layer four projected onto a linear subspace where two axes encode token positions (horizontal axes), and one axis encodes part-of-speech (vertical axis). Projecting from the side visualizes the part-of-speech axis (top right). Projecting from the top down visualizes the token position axes (bottom right).}
    \label{fig:multiple-features1}
\end{figure}
\setlength{\belowcaptionskip}{0.0cm}

We performed LDA on sets of representations corresponding to POS tags in the Universal Dependencies (UD) dataset \citep{nivre-etal-2020-universal}.
Specifically, we mapped language model tokens to the POS tag(s) that they were annotated with anywhere in the UD corpus.\footnote{The UD corpus contained 61 of the 88 languages in both OSCAR and XLM-R.}
Using this mapping from tokens to POS tags, we extracted token representations in each language for each POS tag.
To identify axes separating specific POS tags using LDA, we used a set of 8K token representations for each POS tag, sampled uniformly from all languages with tokens appearing in the UD corpus.
When projecting onto $n$ dimensions, we used LDA over $n+1$ POS tags, resulting in $n$ axes that separated representations for the provided POS tags.\footnote{We omit visualizations applying LDA over all 17 POS tags in the UD dataset because this method resulted in 16 different axes, and the first $n$ axes for visualization generally did not separate all POS tags. In particular, the first three dimensions tended to separate punctuation, numbers, and symbols (e.g. percent signs and emojis), grouping all other POS tags together.}

\textbf{POS axes were language-neutral and stable across layers.}
As with token positions, POS information was encoded largely language-neutrally along POS axes.
As shown in Figure \ref{fig:pos-lda}, when projected onto POS subspaces, representations clustered roughly by POS independently of their input language.
This result aligns with previous work showing that syntactic information aligns in shared linear subspaces across languages \citep{chi-etal-2020-finding}.
Unlike \citet{chi-etal-2020-finding}, we identified low-dimensional subspaces that allow us to visualize representations projected directly onto the subspaces without additional distortion (e.g. unlike $t$-SNE visualizations).
Furthermore, we found that these POS axes were relatively stable across layers one through ten, resulting in similar projections when projecting onto POS axes identified for other layers (see Appendix \ref{app:pos-lda-plots} for additional plots).
This result aligns with the hypothesis that middle layers process higher level information \citep{jawahar-etal-2019-bert,tenney-etal-2019-bert}, but it also suggests that low-level information is still retained along stable axes in these layers.

\setlength{\belowcaptionskip}{-0.3cm}
\begin{figure}[t]
    \centering
    \includegraphics[width=7.5cm]{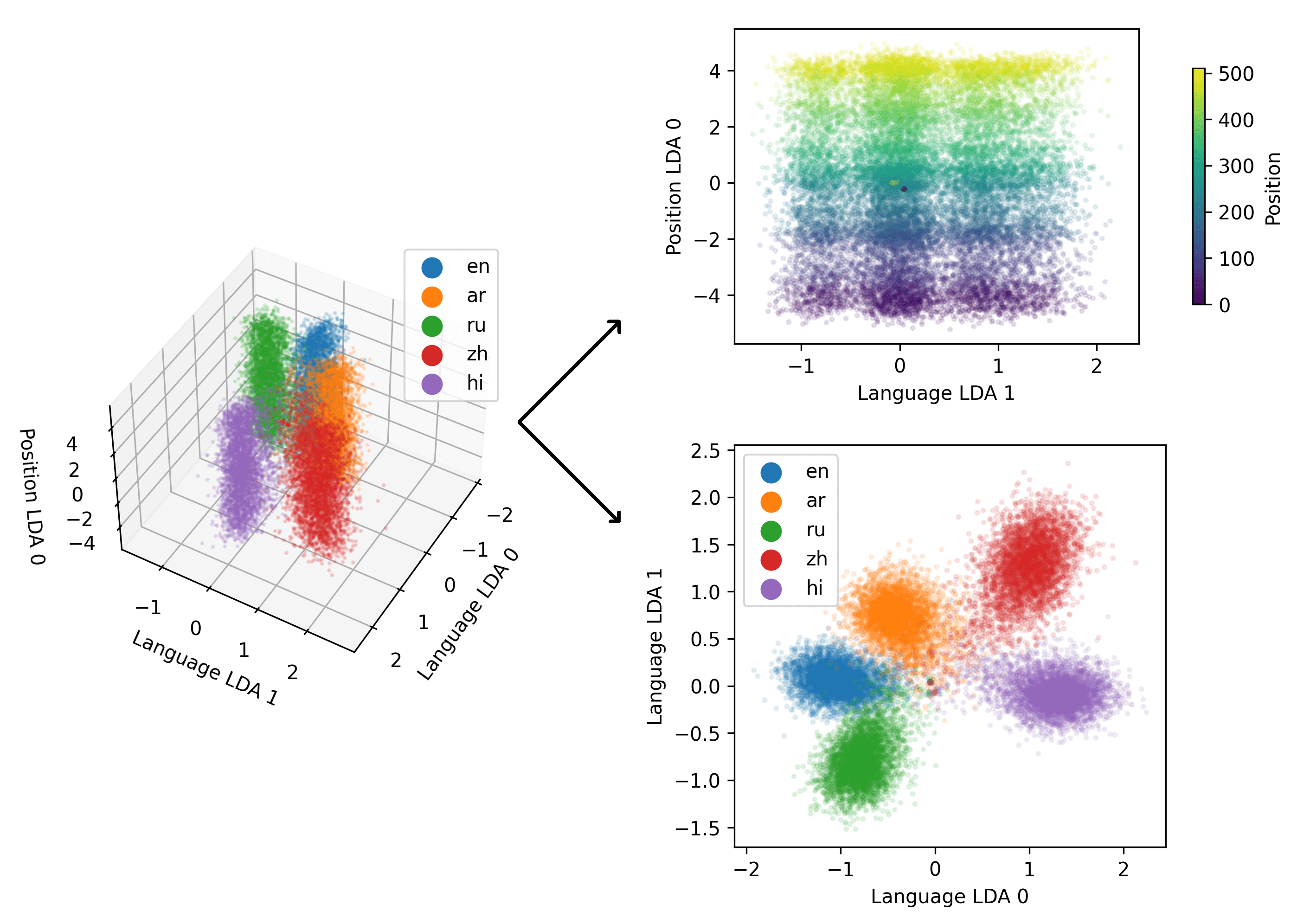}
    \caption{Representations from layer six projected onto a linear subspace where two axes are language-sensitive (horizontal axes), and one axis encodes token positions (vertical axis). Projecting from the side visualizes the language-neutral token position axis (top right). Projecting from the top down visualizes the language-sensitive axes separating languages (bottom right).}
    \label{fig:multiple-features2}
\end{figure}
\setlength{\belowcaptionskip}{0.0cm}

\section{Multilingual structure}
\label{sec:multilingual-structure}
Finally, we synthesize results from previous sections to develop a clearer picture of how multilingual language model representation spaces are structured.
In Section \ref{sec:language-subspaces}, we showed that individual languages occupy affine subspaces that are roughly similar to one another after mean-shifting.
These language subspaces encode information such as token positions and part-of-speech along shared language-neutral axes (Section \ref{sec:language-neutral-axes}).
The subspaces differ primarily along language-sensitive axes (e.g. axes connecting language means) that encode information such as token vocabularies (Section \ref{sec:language-sensitive-axes}).
We note that to the extent to which language subspaces differ along at least one language-sensitive axis, they are esentially non-overlapping in the higher dimensional space.
To be precise, representations in different languages do not directly occupy a shared multilingual subspace; rather, representations must be projected onto specific axes and subspaces to extract language-sensitive and language-neutral representations.
We describe different types of language-sensitive and language-neutral axes in Appendix \ref{app:language-sensitive-neutral}, but we leave a rigorous quantification of the ``language-sensitivity'' of individual features and axes to future work.

Still, as shown in Figures \ref{fig:projections-lang-positions}, \ref{fig:multiple-features1}, and \ref{fig:multiple-features2}, we found that axes encoding different features were often orthogonal to and independent from one another, enabling projections that were minimally influenced by extraneous noise and features along other dimensions.
For example, in Figure \ref{fig:multiple-features1}, the distinction between nouns and verbs was encoded along an axis orthogonal to a token's position in the input sequence, allowing either feature to be extracted by projecting onto its corresponding axes (see additional examples in Appendix \ref{app:multiple-features}).
However, this observation may be because low-dimensional subspaces are likely to be orthogonal to one another in high-dimensional spaces, and we selected features that were likely to be minimally correlated with one another.
Future work might assess how multilingual language models encode more complex linguistic features geometrically, or the extent to which the models' representation spaces can be fully decomposed into orthogonal subspaces encoding different features.
This work has implications for more targeted subspace alignment between languages for better cross-lingual transfer learning; for example, our methods might be used to identify specific axes that should or should not be aligned for specific tasks using existing representation alignment approaches (e.g. \citealp{cao-etal-2020-multilingual, kulshreshtha-etal-2020-cross,zhao-etal-2021-inducing}).

Finally, in Sections \ref{sec:language-lda}, \ref{sec:position-axes}, and \ref{sec:pos-axes}, we presented initial results suggesting that language families, token positions, and part-of-speech are encoded stably across middle and late-middle layers in multilingual language models.
In other words, the transformations in these layers may be primarily language-neutral, and they may retain structures encoding low-level features for later token reconstruction \citep{voita-etal-2019-bottom}.
In these layers, representations might change primarily along axes corresponding to higher level features (e.g. semantics and general reasoning; \citealp{jawahar-etal-2019-bert,tenney-etal-2019-bert}), leaving other axes unchanged.
A more detailed analysis of how specific representational structures are retained across layers is a promising direction for future research.

\section{Conclusion}
In this work, we identified language subspaces and individual language-sensitive and language-neutral axes in the multilingual language model XLM-R.
We assessed these subspaces and axes using a variety of methodologies, including causal effects on language modeling predictions, direct comparisons between subspaces, and low-dimensional visualizations.
Our results suggest that multilingual language models encode features by projecting representations onto orthogonal axes in the representation space, enabling the efficient and simultaneous encoding of a wide variety of signals for downstream tasks and multilingual learning.

\section*{Limitations}
Of course, our work has several limitations.
First, our results were limited by the languages available in XLM-R and the OSCAR corpus \citep{abadji-etal-2021-ungoliant, conneau-etal-2020-unsupervised}.
Our 88 considered languages were skewed substantially towards Indo-European languages; 52 of the 88 languages were in some subfamily of the Indo-European languages (see individual languages in Appendix \ref{app:language-lda-plots}; language families obtained from the Glottolog database; \citealp{hammarstrom-etal-2021-glottolog}).
The 88 languages included only six Austronesian languages, two African languages, and zero Native American languages.
Due to this limited linguistic diversity, our results may overestimate the similarity between subspaces and representations across languages.

Second, even among the considered languages, corpora varied substantially in both size and quality, both in OSCAR and the XLM-R pre-training corpus.
Languages with smaller and less clean corpora are likely to have less-nuanced learned representations and less-representative extracted subspaces in XLM-R.
It is also likely that language-sensitive axes and subspaces encode topic distribution shifts across languages, along with the language families and token vocabularies observed in Section \ref{sec:language-sensitive-axes}.
From this perspective, investigating language-sensitive axes for topic information might allow researchers to quantify how topics differ across languages.

Finally, due to limited computation resources, our experiments were run only on one pre-trained language model, XLM-R.
Different architectures, hyperparameter settings, and parameter initializations could produce different results.
We hope that future work will continue to assess the geometry of multilingual language model representations, covering a wider variety of models and languages.

\section*{Acknowledgements}
We would like to thank Ndapa Nakashole, Leon Bergen, and the UCSD Language and Cognition Lab for helpful discussion and input.
We would also like to thank the anonymous reviewers for valuable feedback.
Tyler Chang is partially supported by the UCSD HDSI graduate fellowship, and Zhuowen Tu is funded by NSF IIS-2127544.

\bibliography{anthology,custom}
\bibliographystyle{acl_natbib}

\appendix

\section{Experimental details}
\label{app:experimental-details}
All of our experiments used the base size XLM-R language model from \citet{conneau-etal-2020-unsupervised} with 270M parameters, implemented in the Huggingface Transformers library \citep{wolf-etal-2020-transformers}.
When extracting representations from XLM-R, we inputted unmodified text from the September 2021 release of the OSCAR corpus of cleaned web text data \citep{abadji-etal-2021-ungoliant}, concatenating lines such that each input sequence contained 512 tokens.
We extracted representations from randomly sampled sequences from the first (up to) 8M such sequences in each language.
Experiments were run on eight Titan Xp GPUs.
Across all reported experiments, total computation time was approximately 17.5 days.
The vast majority of computation time was spent computing evaluation perplexities and in-language token proportions for each projection type, language pair, and layer in Sections \ref{sec:affine-subspaces} and \ref{sec:inducing-vocabulary}.

\section{Subspace distances}

\subsection{Subspace distance metric}
\label{app:distance-metric}
For each language $A$, we obtained a data matrix $\bm{X}_A \in \R^{n \times d}$ of $n$ contextualized token representations in language $A$ from the desired layer in XLM-R.
We used $n=262\textrm{K}$ (512 sequences in the OSCAR corpus), which we found to produce relatively stable subspaces for each language, as quantified by low distances between subspaces computed from different random subsets of representations from language $A$.
Intuitively, the specific random subset of tokens from language $A$ should not impact the computed subspace.
Only two of the 88 selected languages had fewer than 262K tokens in the OSCAR corpus: Javanese ($n=193\textrm{K}$) and Sundanese ($n=70\textrm{K}$).

To represent each language subspace, we considered the covariance matrix $\bm{K}_A \in \R^{d \times d}$ of the original representation dimensions.
If we mean-centered the data matrix $\bm{X}_A$ according to $\bm{\mu}_A \in \R^d$, then $\bm{K}_A$ could be computed as $\bm{X}_A^T \bm{X}_A = \bm{V}_A\bm{\Sigma}_A^2\bm{V}_A^T$ divided by $n-1$, where $\bm{V}_A \in \R^{d \times d}$ were the right singular vectors of $\bm{X}_A$, and $\bm{\Sigma}_A \in \R^{d \times d}$ was the diagonal matrix of $d$ singular values of $\bm{X}_A$ (computed using singular value decomposition).
Geometrically, $\bm{K}_A$ can be interpreted as an ellipsoid with axis directions defined by the (orthonormal) columns of $\bm{V}_A$ and axis scales defined by the corresponding variances in those directions ($\bm{\Sigma}_A^2$ scaled down by $n-1$).
This ellipsoid is the image of the unit sphere under the linear transformation $\bm{K}_A$.
In this way, $\bm{K}_A$ captures the variance in different directions for representations from language $A$.

To compute the distance between two covariance matrices $\bm{K}_A$ and $\bm{K}_B$, we used the distance metric from \citet{bonnabel-sepulchre-2009-riemannian} between positive definite matrices, written out as Equation \ref{eq:distance} in Section \ref{sec:subspace-distances}.
Motivating this metric, \citet{bonnabel-sepulchre-2009-riemannian} describe a natural Riemannian metric on the space of positive definite matrices.
A Riemannian metric defines an inner product over the tangent space at each point in the manifold (in this case, at each positive definite matrix).
These inner products can be used to define the lengths of paths (geodesics) in the manifold.
Then, the distance between any two positive definite matrices can be defined by the length of the shortest geodesic curve connecting them.
The resulting metric is both symmetric and invariant to linear transformations.
Further theoretical properties of the metric are described in \citet{bonnabel-sepulchre-2009-riemannian}.

\subsection{Rotation and scaling comparisons}
\label{app:rotations-scalings}
In Section \ref{sec:subspace-distances}, we computed analogous rotation degrees and scaling multipliers for language subspace distances in each layer of XLM-R.
To do this, for each layer of XLM-R, we first computed the mean distance from each language subspace to itself before and after rotation by $\theta$ degrees or scaling by multiplier $\gamma$ along each axis.
We computed this mean over all languages and over 16 random rotations or scalings, considering $\theta \in [0, 1, ..., 90]$ and $\gamma \in [1.00, 1.01, ..., 4.00]$.
Random rotations were computed by sampling $d//2$ independent planes of rotation (by generating a random orthonormal basis and pairing together consecutive dimensions) and rotating the principal axes (columns of $\bm{V}_A \in \R^{d \times d}$ as computed in Appendix \ref{app:distance-metric}) by $\pm \theta$ degrees in each plane.
Random scalings were computed by multiplying or dividing the variance along each principal axis by $\gamma^2$, because scaling representations by $\gamma$ along each axis would scale the corresponding variances by $\gamma^2$.
As expected, we found that the mean distance between transformed subspaces increased monotonically with respect to both rotation degree and scaling multiplier.

Then, given a true distance value between two language subspaces, we computed the equivalent rotation degree (and scaling multiplier) by identifying the lowest degree (or scaling multiplier) that would result in an equal or greater distance value.
For example, given a true distance value of $30.0$, we identified the lowest rotation degree that resulted in a mean distance of at least $30.0$ when language subspaces were rotated individually.
Essentially, we sought to identify the location (in rotation degrees or scaling multipliers) along the monotonically increasing distance curve where each true distance value would fall.
We computed this rotation degree and scaling multiplier for each language pair's distance in each XLM-R layer, with results described in Section \ref{sec:subspace-distances}.

\section{Language-sensitive vs. language-neutral axes}
\label{app:language-sensitive-neutral}
Here, we consider in more detail what it means for an axis or subspace to be language-sensitive or language-neutral.
Broadly, as in Section \ref{sec:language-sensitive-axes}, we call a subspace language-sensitive if when a representation is projected onto the subspace, it has high mutual information with its input language identity.
For example, if representations from two languages have different distributions when projected onto some subspace, then the projected representations implicitly contain information about their input languages.
Then, we call the subspace language-sensitive; conversely, if the mutual information is low, we call the subspace language-neutral.

Notably, when considering linear axes and subspaces, we can consider the means and variances of projected representations from different languages.
Specifically, we can have the following language-neutral axes:
\begin{enumerate}[label=(\roman*)]
\item Equal means, low variance in each language: due to low variance, these axes are not likely to contain a significant amount of relevant information within or across languages, although variance is not a perfect indicator for downstream importance. \citet{rajaee-pilehvar-2022-isotropy} suggest that there may be relatively few ``degenerate'' axes of this type in multilingual language models because languages tend to be ``degenerate'' along different axes, but they only considered axes in the original basis of the representation space.
\item Equal means, similarly high variance in each language: if projected representations from different languages have similar distributions (on top of similar means and variances) along these axes, then the axes are indeed language-neutral.
This can be observed qualitatively by viewing distributions of projected representations in different languages (e.g. Figures \ref{fig:position-lda} and \ref{fig:pos-lda} in Section \ref{sec:language-neutral-axes}).
\end{enumerate}
And the following language-sensitive axes:
\begin{enumerate}[label=(\roman*), resume]
\item High variance in $A$, low variance in $B$: these axes may represent features that vary in language $A$ but not in language $B$.
Previous work has found evidence for these types of axes in multilingual language models (low cosine similarities along certain axes only in specific languages; \citealp{rajaee-pilehvar-2022-isotropy}), characterized as ``degenerate'' axes only in some languages.
In our work (Section \ref{sec:inducing-vocabulary}), we found that setting representations from language $A$ equal to $\bm{\mu}_B$ along axes with low variance in language $B$ induced more token predictions in language $B$ than shifting by $\bm{\mu}_B - \bm{\mu}_A$ alone.
In other words, there was variance in language $A$ (centered around $\bm{\mu}_A$) that may have been interpreted as noise (resulting in tokens outside language $B$) if simply shifted to be centered around $\bm{\mu}_B$; setting the representations equal to $\bm{\mu}_B$ along these axes was sufficient to remove this noise.
These axes may have been encoding specific tokens or token features in language $A$, less relevant to language $B$ (high variance in $A$, low variance in $B$).

In these cases, $\bm{\mu}_B$ may be close to or far from $\bm{\mu}_A$, depending on whether language $B$ has some fixed analogous feature value for the feature that varies in language $A$.
Because there is little to no variance in language $B$ along these axes, $\bm{\mu}_B$ essentially serves as a fixed bias term for language $B$ representations.
\item \label{item:unequal-means-high-variances} Unequal means, similarly high variance in each language: visually, these axes might represent languages as similar distributions shifted from one another (e.g. Figure \ref{fig:lang-lda}, depicting language clusters). These shifts might reflect true differences between languages in the distributions of particular features, imperfect alignment between languages, or entirely different features for different languages (see Appendix \ref{app:encoding-features} below).
\item Unequal means, low variance in each language: due to the low variance within each language, these axes would represent different languages roughly as points, potentially encoding any of the wide variety of features that vary across languages (although with little variance within each language).
However, there is little evidence that these axes exist in multilingual language models.
Even axes identified using LDA between languages (designed to maximize variance between languages and minimize variance within languages) appeared closer to type \ref{item:unequal-means-high-variances} above.
\end{enumerate}
Future work may consider more rigorous quantifications of the language-sensitivity of individual axes and subspaces in multilingual language models, possibly quantifying the relative dimensionalities of the language-neutral and language-sensitive subspaces.
Future work may also consider nonlinear subspaces in the models; as demonstrated by the spirals, toruses, and curves in Section \ref{sec:position-axes}, some features are represented nonlinearly in multilingual language models.

\subsection{Extracting features}
\label{app:encoding-features}
However, the interpretations above primarily consider representation spaces as isolated units; they do not consider how the models extract information from the spaces.
For example, even if a subspace appears language-neutral, features might be extracted from the projected representations in language-sensitive ways, and the subspace axes might thus have different effects on downstream processing for different languages.
For example, representation distributions in languages $A$ and $B$ might look similar along some ``language-neutral'' axis, but the distributions might encode different features for downstream processing in the two languages.
We attempt to avoid this confound in our work by explicitly hypothesizing interpretable language-neutral features (e.g. token positions or part-of-speech in Section \ref{sec:language-neutral-axes}) that might be encoded along particular language-neutral axes.
We can then assess whether individual axes encode specific features across languages, although causal effects on downstream model predictions need to be verified independently (e.g. Section \ref{sec:inducing-vocabulary}).

More broadly, given any language-sensitive or language-neutral axis, we cannot distinguish solely on the basis of representation distributions in each language whether the axis encodes the same feature or the same type of information across languages.
By grounding our analyses in interpretable features and downstream consequences, we can begin to identify how individual axes affect model processing in language-sensitive or language-neutral ways.

\section{Additional plots}
In this section, we include additional visualizations projecting representations onto language-sensitive and language-neutral axes.
In all visualizations, we use the global representation mean $\bm{\mu}_{\textrm{global}}$ across languages as our origin along each axis.
The choice of mean only shifts projected representations by a constant vector, not affecting visualizations.
We orthogonalize and normalize our axes prior to projection, ensuring that representations are projected directly onto the corresponding subspaces without stretching or distortion.

\subsection{Language-sensitive axes}
\label{app:language-lda-plots}
\begin{figure*}[!t]
    \centering
    \includegraphics[width=16cm]{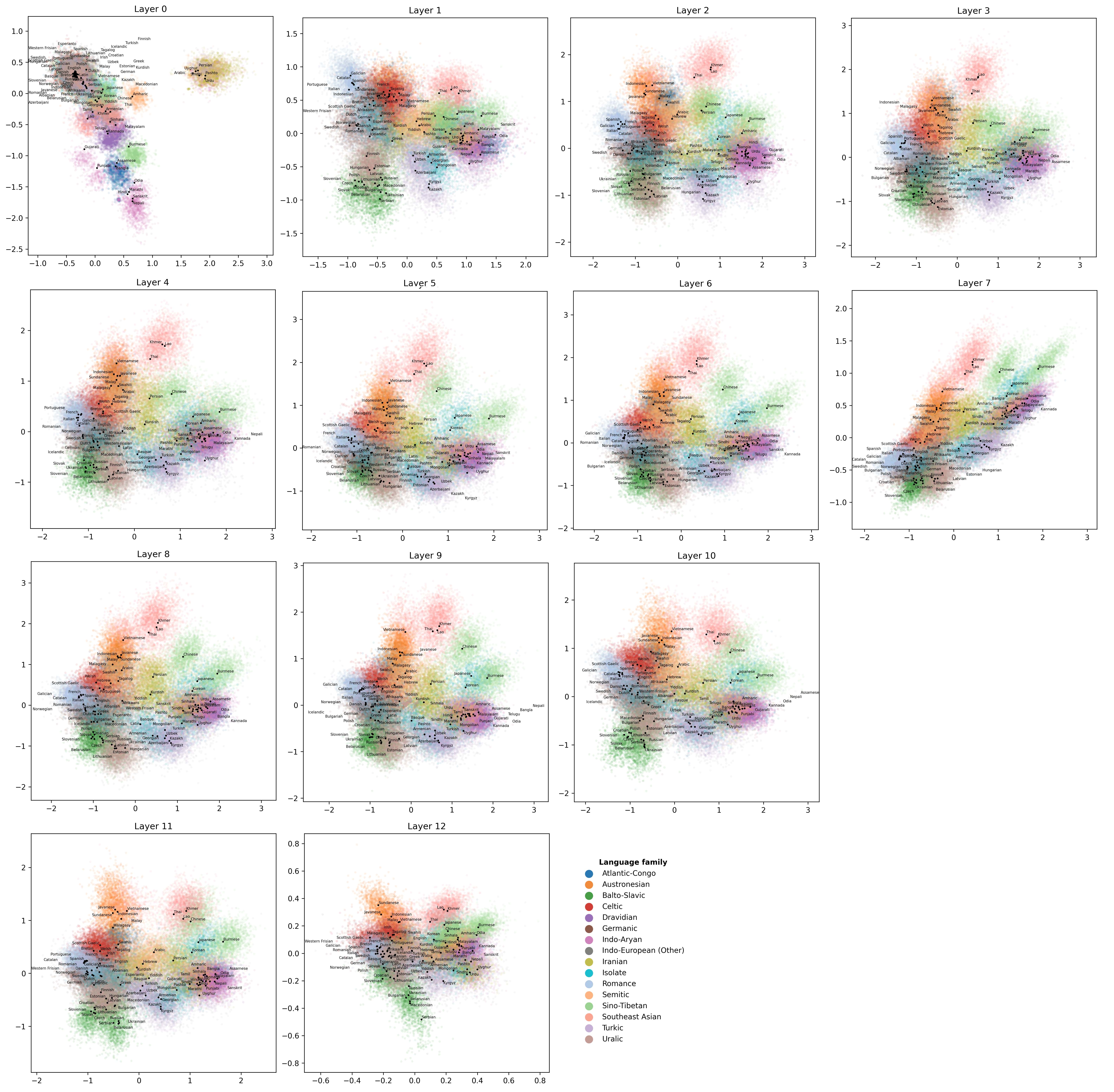}
    \caption{Representations in each layer projected onto the first two LDA axes that separate languages in that layer.
    Points indicate language means.
    Language families were obtained from the Glottolog database  \citep{hammarstrom-etal-2021-glottolog}, with varying levels of granularity depending on the number of XLM-R languages in each family (e.g. breaking down the Indo-European languages).
    We grouped together the two Southeast Asian language families in XLM-R: Austroasiatic (Vietnamese and Khmer) and Tai-Kadai (Thai and Lao).
    We identified language isolates as languages that were the only XLM-R language in their family (Basque, Esperanto, Georgian, Japanese, Korean, and Mongolian). Best viewed on a computer for magnification.}
    \label{appfig:language-ldas-means}
\end{figure*}

\begin{figure*}[t]
    \centering
    \includegraphics[width=16cm]{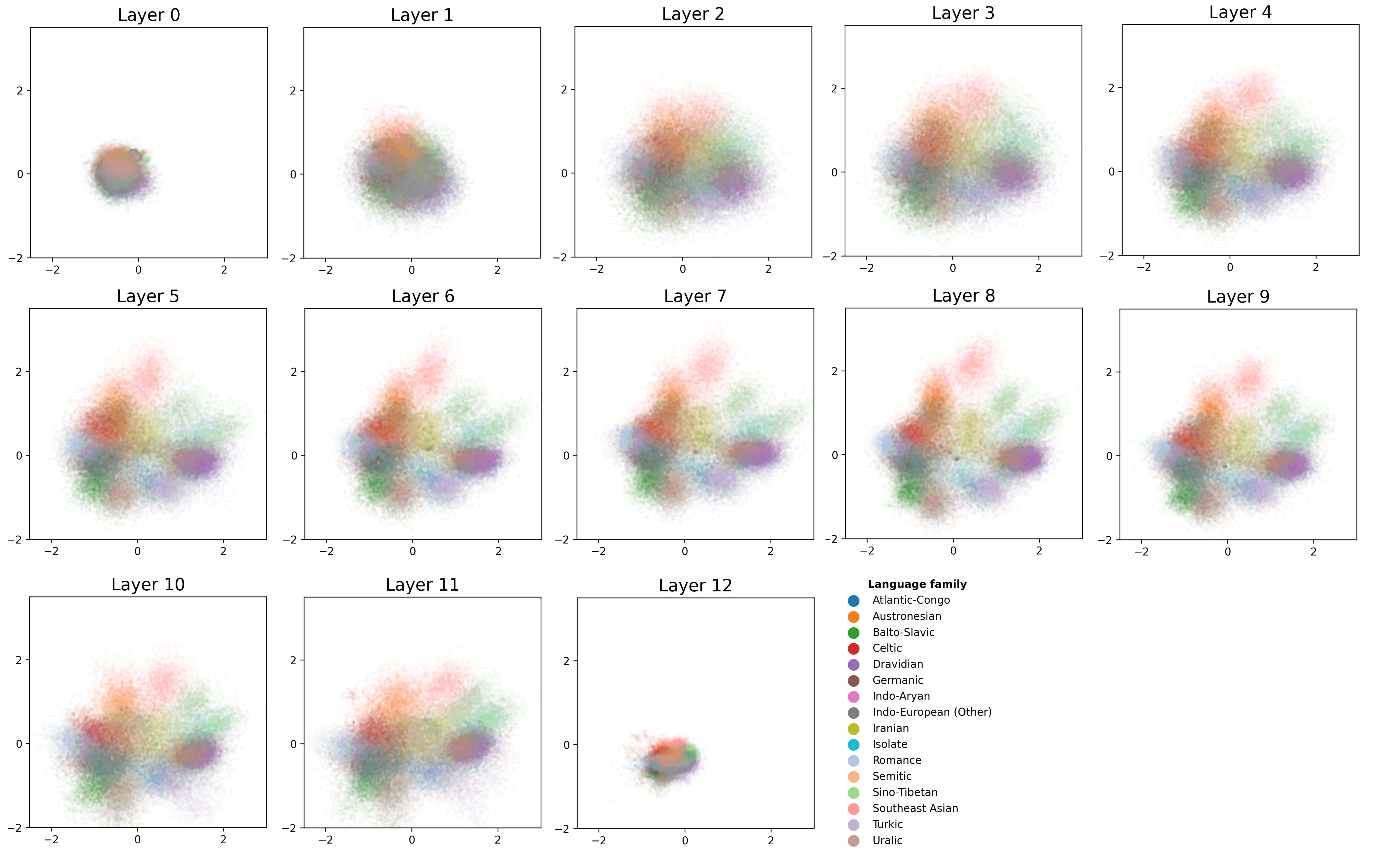}
    \caption{Representations from each layer projected onto the first two LDA axes that separate languages in layer eight. The structure of representations along these axes was relatively stable across middle layers.}
    \label{appfig:language-ldas-crosslayer}
\end{figure*}

We projected representations onto the first two LDA axes that separated languages in each layer (Figure \ref{appfig:language-ldas-means}), including all 88 language means. In Figure \ref{appfig:language-ldas-crosslayer}, we show representations from each layer projected onto the LDA axes identified specifically for layer eight, showing that these axes resulted in similar projections across all middle layers.

\setlength{\belowcaptionskip}{0.5cm}
\setlength{\belowcaptionskip}{0.0cm}

\subsection{Token positions}
\label{app:position-lda-plots}
\begin{figure*}[t]
    \centering
    \includegraphics[width=16cm]{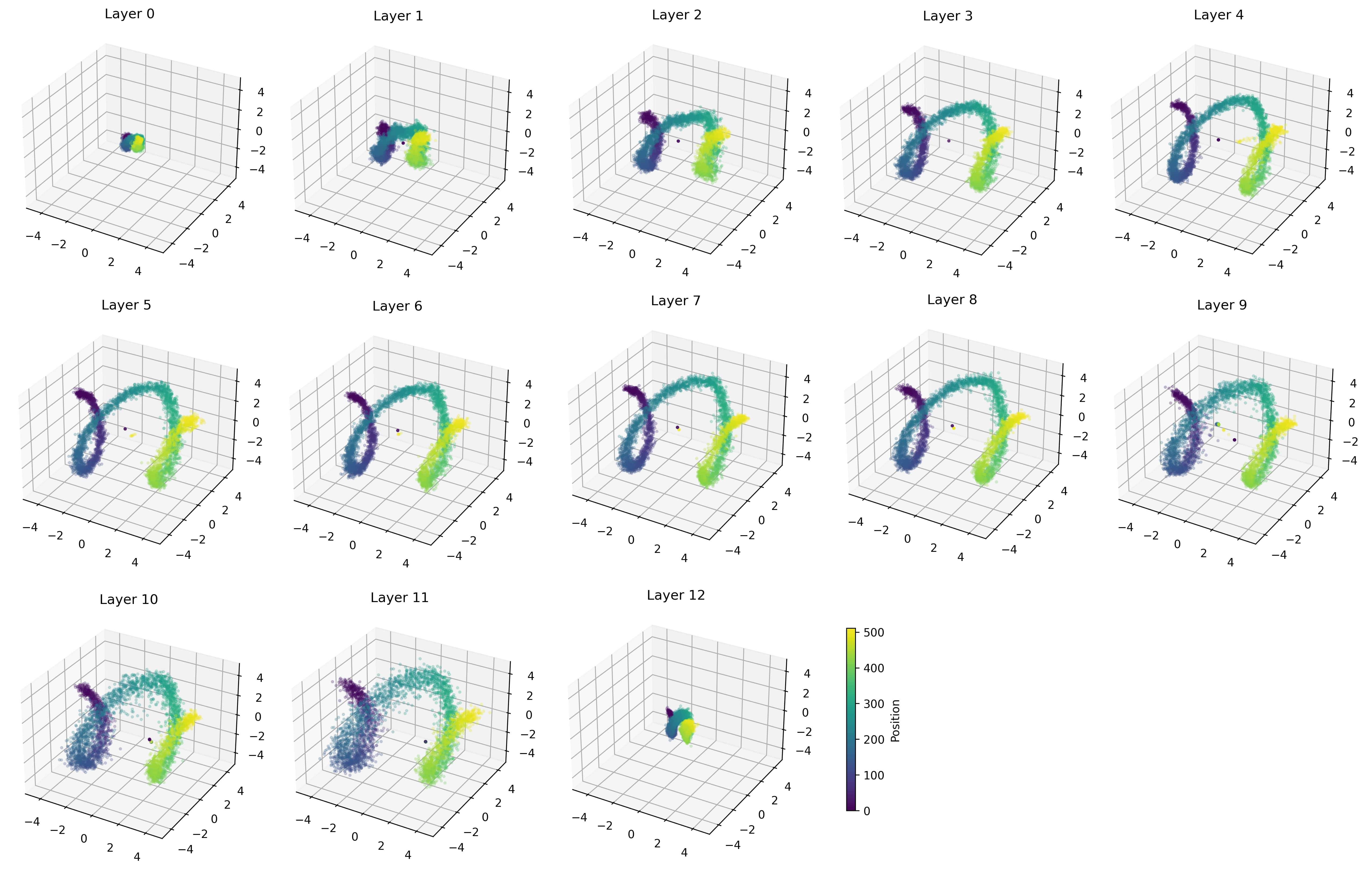}
    \caption{Representations from each layer projected onto the first three LDA axes that separate token positions in layer eight. Token position information was encoded in stable structures along stable axes throughout middle layers.}
    \label{appfig:position-ldas-crosslayer}
\end{figure*}

To demonstrate that token position information is encoded stably across layers, we projected representations from each layer onto the first three LDA axes that separated token positions in layer eight, obtaining similar plots for layers two through eleven (Figure \ref{appfig:position-ldas-crosslayer}).

\subsection{Part-of-speech}
\label{app:pos-lda-plots}
\begin{figure*}[t]
    \centering
    \includegraphics[width=16cm]{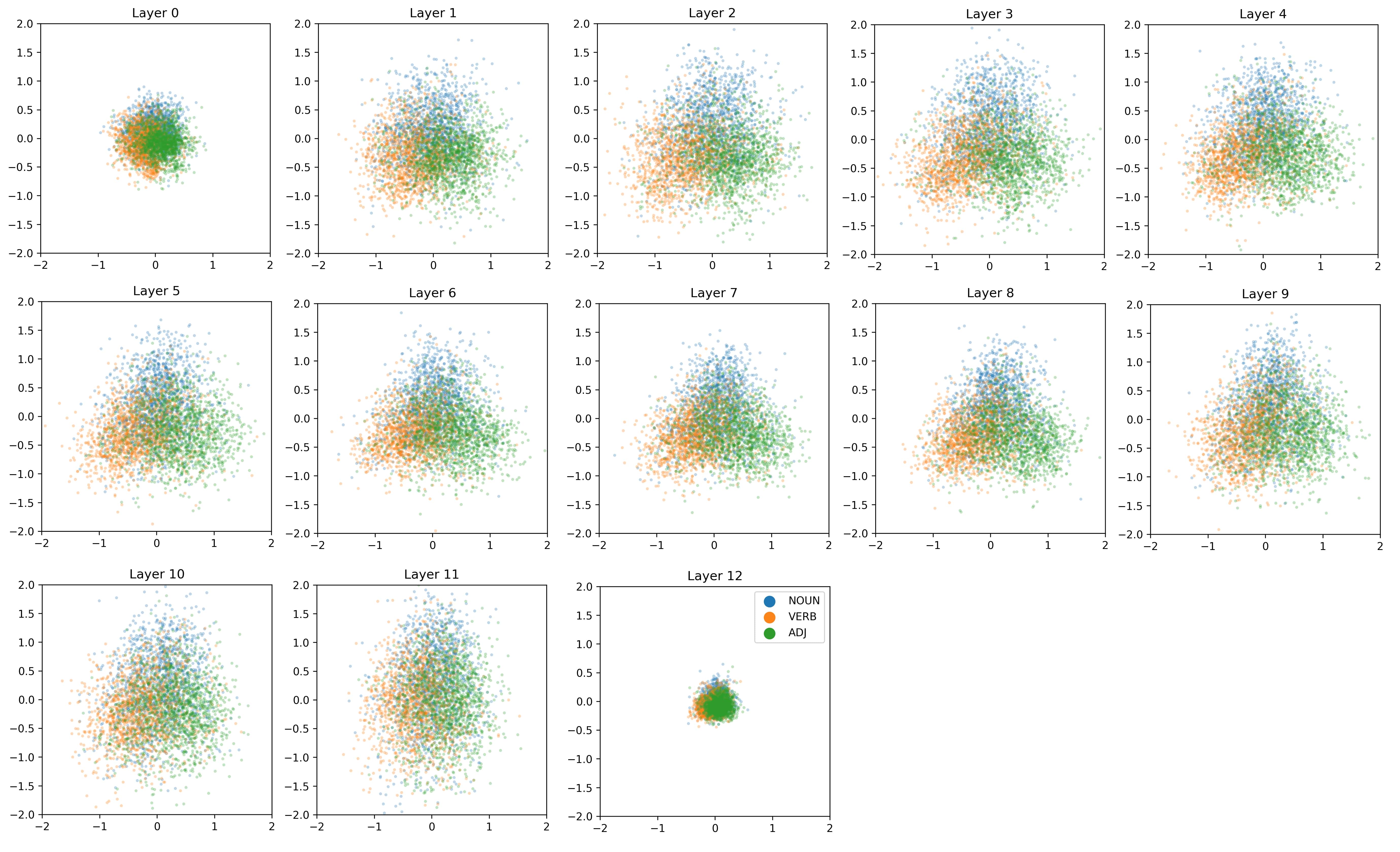}
    \caption{Representations from each layer projected onto the LDA axes that separate nouns, verbs, and adjectives in layer eight. The axes encoded this information stably throughout middle layers. The axes were computed only from LDA in layer eight, but they encoded the same information in other layers as well.}
    \label{appfig:pos-ldas-crosslayer}
\end{figure*}

\begin{figure}[t]
    \centering
    \includegraphics[width=7.5cm]{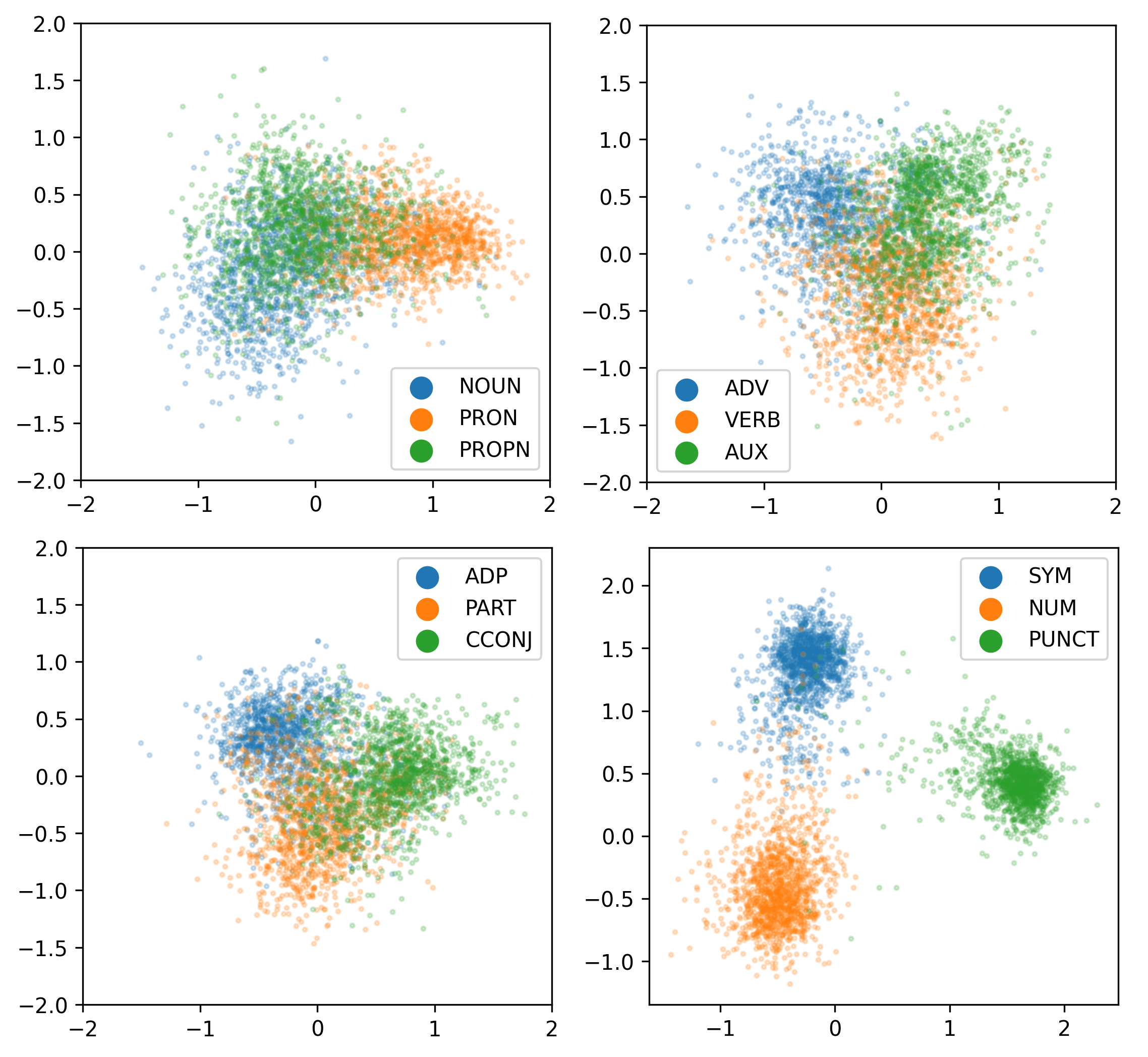}
    \caption{Representations from layer four projected onto the LDA axes that separate the parts-of-speech shown in each figure. Top left: nouns, pronouns, and proper nouns. Top right: adverbs, verbs, and auxiliary verbs. Bottom left: adpositions, particles, and coordinating conjunctions. Bottom right: symbols, numerals, and punctuation.}
    \label{appfig:pos-ldas}
\end{figure}

As shown in Figure \ref{appfig:pos-ldas}, LDA was able to identify axes that separated parts-of-speech (POS).
Furthermore, these axes were stable across the majority of layers; we obtained similar plots when projecting representations from layers one through ten onto POS axes identified for layer eight (Figure \ref{appfig:pos-ldas-crosslayer}).

\subsection{Multiple features}
\label{app:multiple-features}
\begin{figure}[t]
    \centering
    \includegraphics[width=7.5cm]{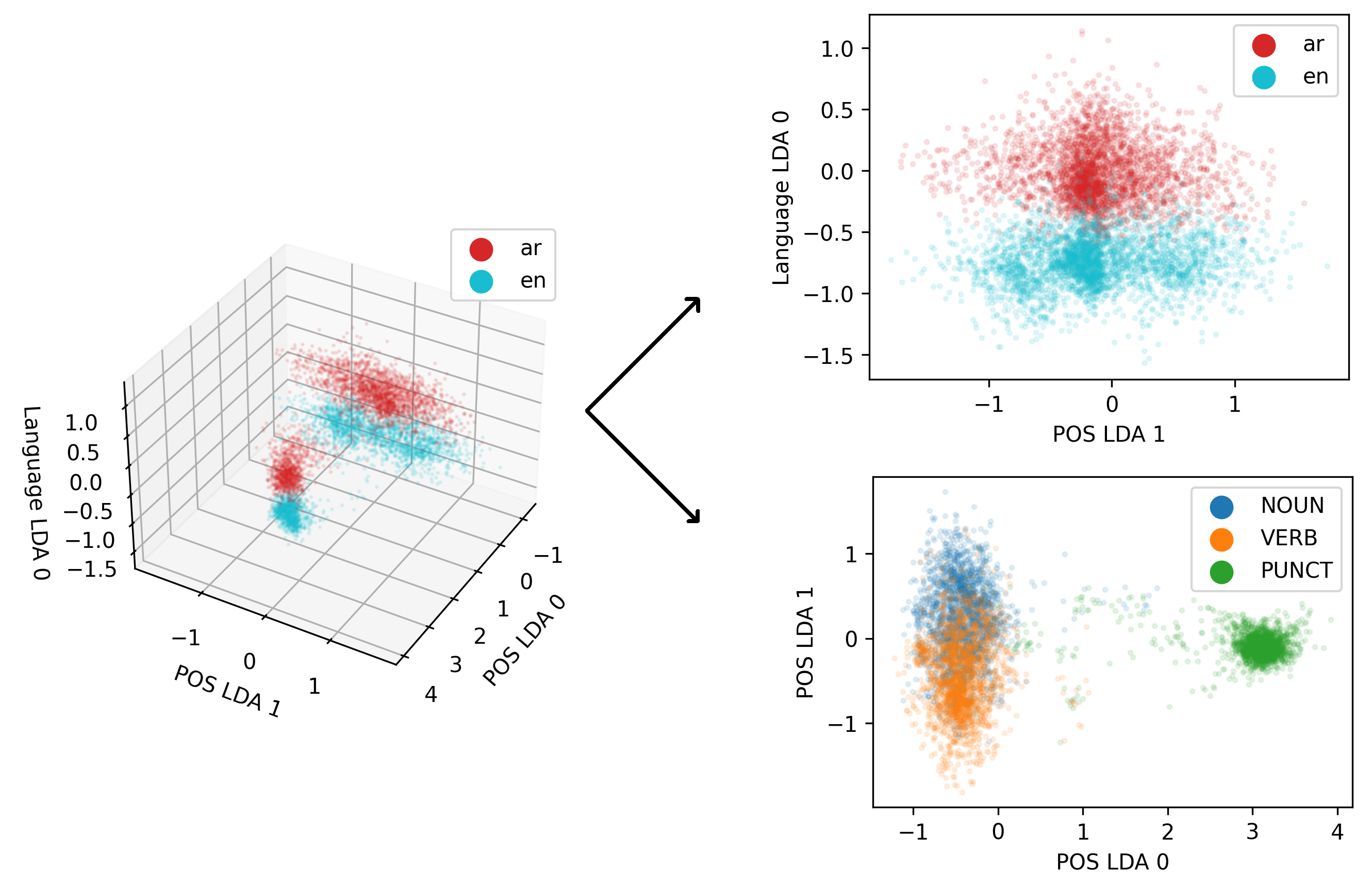}
    \caption{Representations from layer two projected onto a linear subspace where two axes encode part-of-speech (horizontal axes), and one axis is language-sensitive (vertical axis). Projecting from the side visualizes the language-sensitive axis (top right). Projecting from the top down visualizes the language-neutral part-of-speech axes (bottom right).}
    \label{appfig:multiple-features1}
\end{figure}

\begin{figure}[t]
    \centering
    \includegraphics[width=7.5cm]{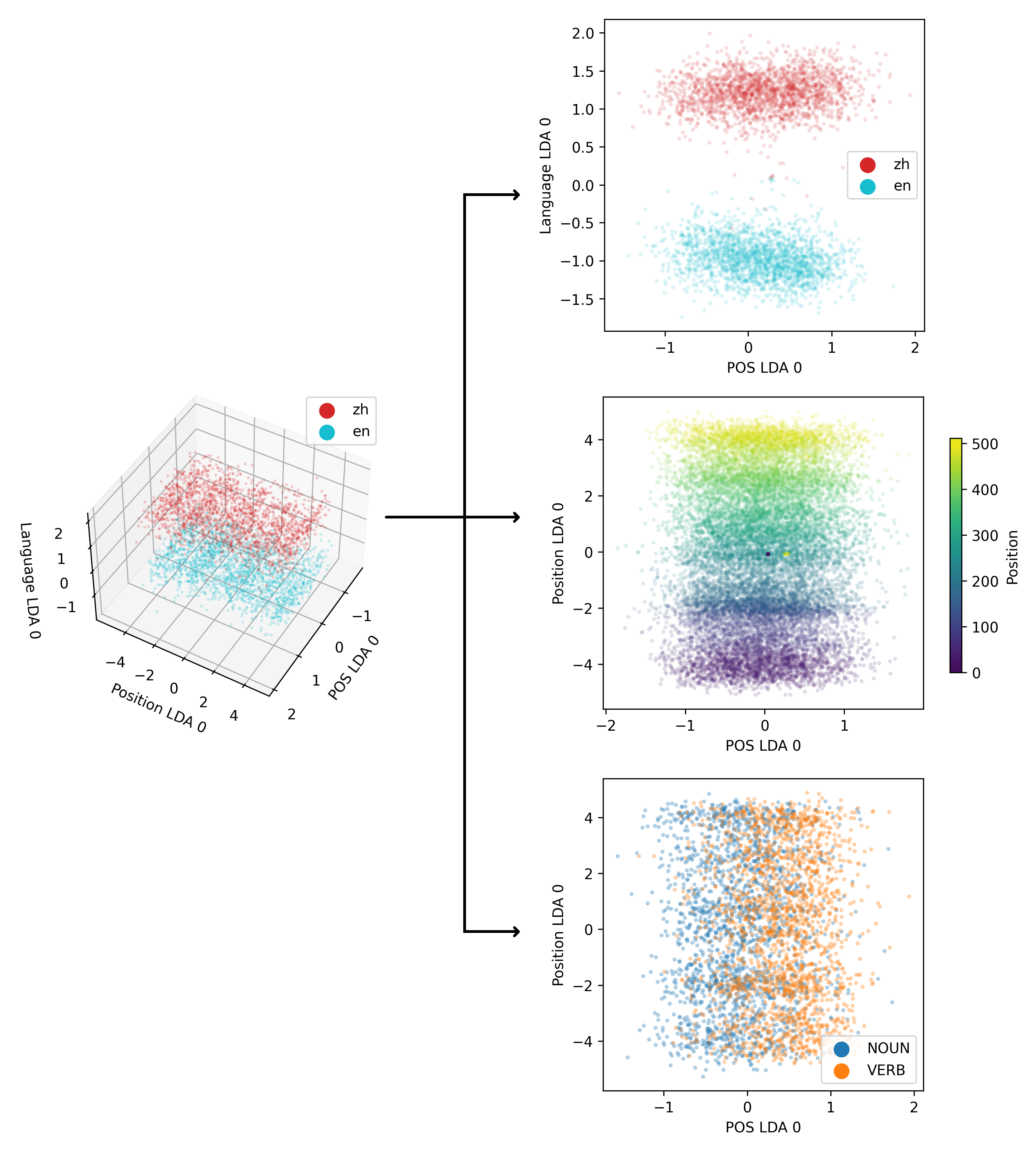}
    \caption{Representations from layer eight projected onto a linear subspace where one axis encodes part-of-speech (first horizontal axis), one axis encodes token positions (second horizontal axis), and one axis is language-sensitive (vertical axis). Projecting from the side visualizes the language-sensitive axis separating languages.
    Projecting from the top down visualizes the language-neutral token position axis (middle right) and part-of-speech axis (bottom right).}
    \label{appfig:multiple-features2}
\end{figure}

Finally, we projected representations onto subspaces where different axes encoded different features (e.g. token positions, part-of-speech, or input language).
In Figures \ref{appfig:multiple-features1} and \ref{appfig:multiple-features2} (along with Figures \ref{fig:projections-lang-positions}, \ref{fig:multiple-features1}, and \ref{fig:multiple-features2} in the main text), projecting onto different axes visualizes different features.

\end{document}